%% file: main.tex
\documentclass[journal]{IEEEtran}

\IEEEoverridecommandlockouts                              

\usepackage{graphics} 
\usepackage[pdftex]{graphicx}
\graphicspath{{./Figures/,./}}
\DeclareGraphicsExtensions{.pdf,.jpeg,.png,.eps,.tif}
\usepackage{epsfig} 
\usepackage{amsmath} 
\usepackage{amssymb}  
\usepackage{bbold}
\usepackage[export]{adjustbox}
\usepackage[]{subfig}
\usepackage{lipsum}
\usepackage{comment}
\usepackage{tabularx,booktabs}
\usepackage{todonotes}
\usepackage{textcomp}
\usepackage{algorithmic}
\newcommand{\R}{\mathbb{R}}

\usepackage{cite}
\usepackage{bm}
\usepackage{soul}
\usepackage{hyperref}
\newcommand{\ra}[1]{\renewcommand{\arraystretch}{#1}}

\usepackage{algorithm}

\makeatletter
  
\newenvironment{formulation}[1][htb]{%
    \let\c@algorithm\c@protocol
    \renewcommand{\ALG@name}{Formulation}
    \begin{algorithm}[#1]%
  }{\end{algorithm}}
\makeatother

\newcommand{\vg}[1]{\bm{#1}}
\renewcommand{\v}[1]{\mathbf{#1}}
\newtheorem{remark}{Remark}

\usepackage{color}

\newcommand*{\revised}{\textcolor{black}}
\newcommand*{\revisedtwo}{\textcolor{black}}

\usepackage[font=footnotesize]{caption}

\makeatletter
\newcommand\mypicture[2][\textwidth]{%
  \setbox0\hbox{\includegraphics[width=\textwidth]{#2}}
  \ifdim\ht0>\dimexpr\pagegoal-\pagetotal\relax
    \@latex@warning{This picture might be oddly placed}\fi
  \begin{strip}
    \includegraphics[width=#1]{#2}
  \end{strip}
}

\usepackage{tikz,xcolor,hyperref}

\definecolor{lime}{HTML}{A6CE39}
\DeclareRobustCommand{\orcidicon}{%
	\begin{tikzpicture}
	\draw[lime, fill=lime] (0,0) 
	circle [radius=0.16] 
	node[white] {{\fontfamily{qag}\selectfont \tiny ID}};
	\draw[white, fill=white] (-0.0625,0.095) 
	circle [radius=0.007];
	\end{tikzpicture}
	\hspace{-2mm}
}

\foreach \x in {A, ..., Z}{%
	\expandafter\xdef\csname orcid\x\endcsname{\noexpand\href{https://orcid.org/\csname orcidauthor\x\endcsname}{\noexpand\orcidicon}}
}

\makeatother

\begin{document}

\title{\LARGE \bf
Real-Time Deformable-Contact-Aware Model Predictive Control for Force-Modulated Manipulation
{\footnotesize}

}


\author{
\large Lasitha Wijayarathne$^{1*}$,~\IEEEmembership{Student Member,~IEEE,}\orcidA{} Ziyi Zhou$^{2*}$,~\IEEEmembership{Student Member,~IEEE,}\orcidB{} \\
Ye Zhao$^{2 \dagger}$,~\IEEEmembership{Senior Member,~IEEE,}\orcidD{} and  Frank L. Hammond III$^{1 \dagger}$,~\IEEEmembership{Member,~IEEE}\orcidC{}
\thanks{$^{1}$Lasitha Wijayarathne and Frank L. Hammond III are with the Adaptive Robotic Manipulation Laboratory in the Woodruff School of Mechanical Engineering and the Coulter Department of Biomedical Engineering, Georgia Institute of Technology. {\tt\small frank.hammond@me.gatech.edu.}}%
\thanks{$^{2}$Ziyi Zhou, and Ye Zhao are with The Laboratory of Intelligent Decision and Autonomous Robots at Woodruff School of Mechanical Engineering, Georgia Institute of Technology. {\tt\small ye.zhao@me.gatech.edu.}}
\thanks{*The first two authors contributed equally to this work.}
\thanks{$\dagger$Co-corresponding authors.}
}
\maketitle 

\begin{abstract}
Force modulation of robotic manipulators has been extensively studied for several decades. However, it is not yet commonly used in safety-critical applications due to a lack of accurate interaction contact modeling and weak performance guarantees - a large proportion of them concerning the modulation of interaction forces. This study presents a high-level framework for simultaneous trajectory optimization and force control of the interaction between a manipulator and soft environments, which is prone to external disturbances. Sliding friction and normal contact force are taken into account. The dynamics of the soft contact model and the manipulator are simultaneously incorporated in a trajectory optimizer to generate desired motion and force profiles. A constrained optimization framework based on Alternative Direction Method of Multipliers (ADMM) has been employed to efficiently generate real-time optimal control inputs and high-dimensional state trajectories in a Model Predictive Control fashion. Experimental validation of the model performance is conducted on a soft substrate with known material properties using a Cartesian space force control mode. Results show a comparison of ground truth and real-time model-based contact force and motion tracking for multiple Cartesian motions in the valid range of the friction model. It is shown that a contact model-based motion planner can compensate for frictional forces and motion disturbances and improve the overall motion and force tracking accuracy. The proposed high-level planner has the potential to facilitate the automation of medical tasks involving the manipulation of compliant, delicate, and deformable tissues.
\end{abstract}

\input{introduction}
\input{related_work}
\input{methods_contact}

\input{methods_admm}

\input{methods_mpc}

\input{experiments}
\input{conclusion}





\input{appendix}
\bibliographystyle{ieeetr}
\bibliography{refs}

\begin{IEEEbiography}[{\includegraphics[width=1in,height=1.25in,clip,keepaspectratio]{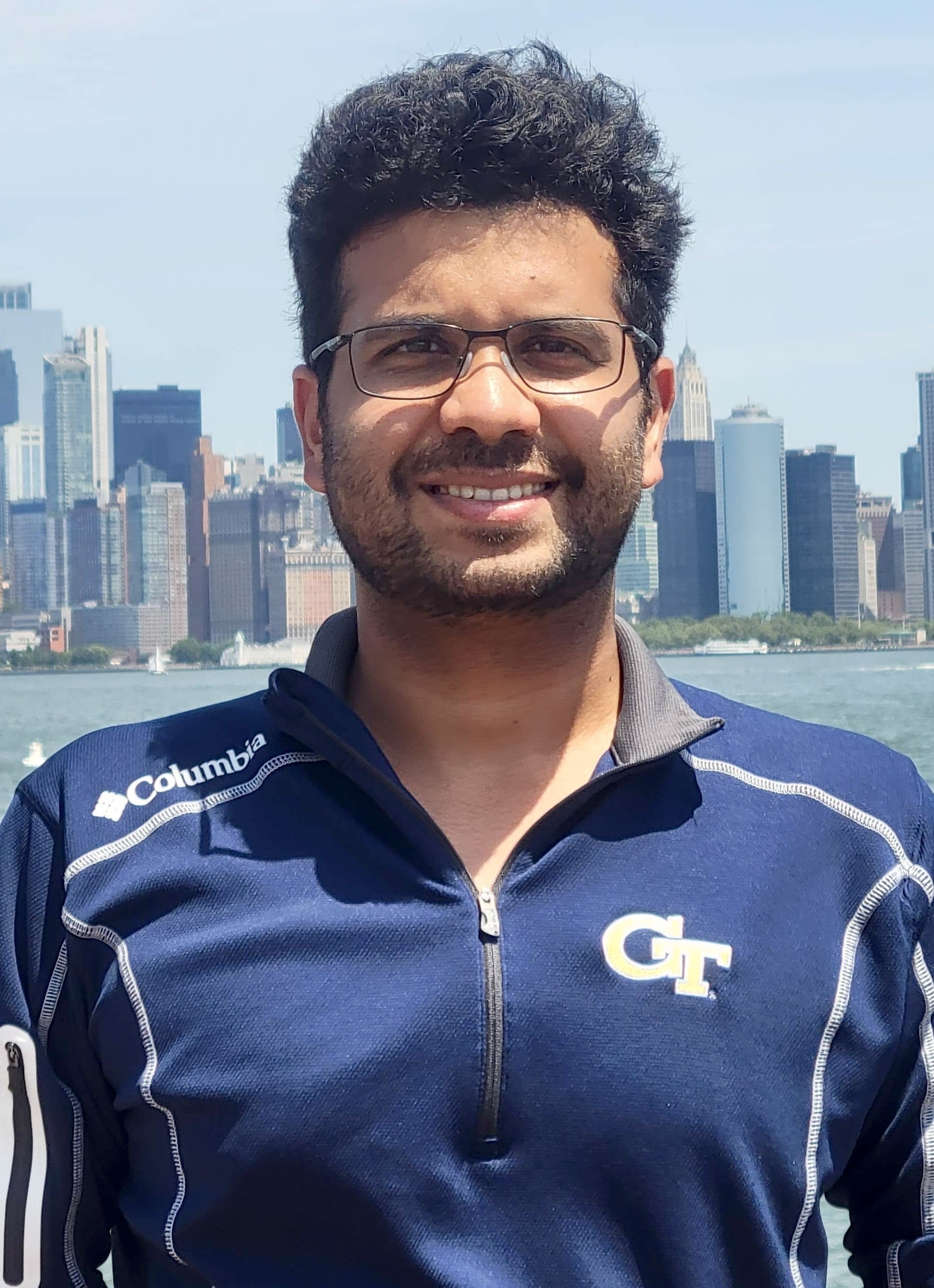}}]{Lasitha Wijayarathne}
Lasitha Wijayarathne received B.S. and M.S. degrees in mechanical engineering from Georgia Institute of Technology, Atlanta, GA, USA. in 2015 and 2019. He is currently a Ph.D. candidate in robotics at Georgia Institute of Technology. His research interests centers around robot design optimization, manipulation, contact-rich trajectory planning on robotic systems.
\end{IEEEbiography}
\vspace{-0.45in}
\begin{IEEEbiography}[{\includegraphics[width=1in,height=1.25in,clip,keepaspectratio]{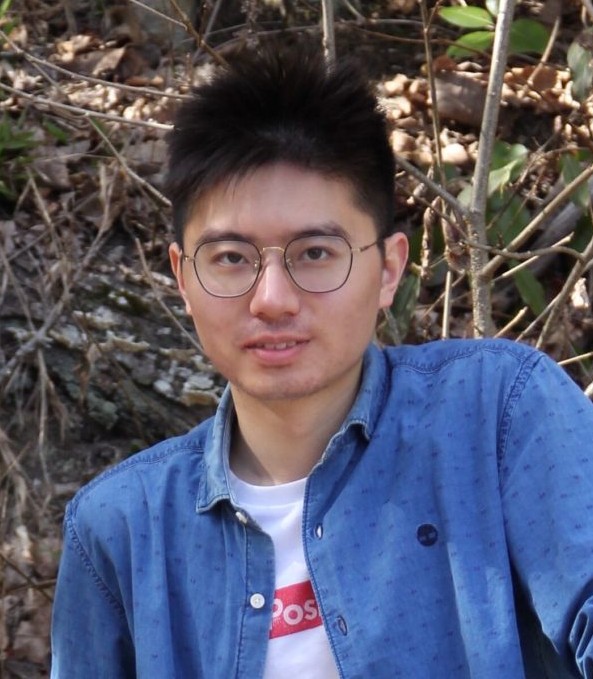}}]{Ziyi Zhou}
Ziyi Zhou received the B.S. degree in automation from Northeastern University, Shenyang, China in 2018 and the M.S. degree in electrical and computer engineering from Georgia Institute of Technology in 2021. He is currently pursuing a Ph.D. degree in electrical and computer engineering at Georgia Institute of Technology with a focus on robotics. His research interests center around contact-rich trajectory optimization and task planning applied to legged robots.
\end{IEEEbiography}

\vspace{-0.55in}
\begin{IEEEbiography}[{\includegraphics[width=1in,height=1.25in,clip,keepaspectratio]{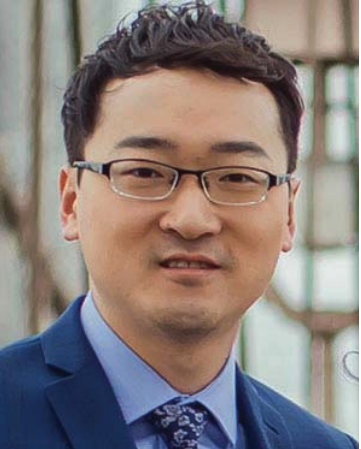}}]{Ye Zhao}
Ye Zhao (Member, IEEE) received the Ph.D. degree in mechanical engineering from The University of Texas at Austin, Austin, TX, USA, in 2016. He was a Post-Doctoral Fellow with the John A. Paulson School of Engineering and Applied Sciences, Harvard University, Cambridge, MA, USA. He is currently an Assistant Professor with the George W. Woodruff School of Mechanical Engineering, Georgia Institute of Technology, Atlanta, GA, USA. His research interests include robust task and motion planning, contact-rich trajectory optimization, legged locomotion and manipulation.
\\ Dr. Zhao serves as an Associate Editor of IEEE Robotics and Automation Letters (RA-L) and IEEE Control Systems Letters (L-CSS). He is a Co-Chair of the IEEE Robotics and Automation Society (RAS) Technical Committee on Whole-Body Control and was a Co-Chair of the IEEE RAS Student Activities Committee and an ICT Chair of the 2018 IEEE/RSJ International Conference on Intelligent Robots and Systems.
\end{IEEEbiography}

\vspace{-0.45in}
\begin{IEEEbiography}[{\includegraphics[width=1in,height=1.25in,clip,keepaspectratio]{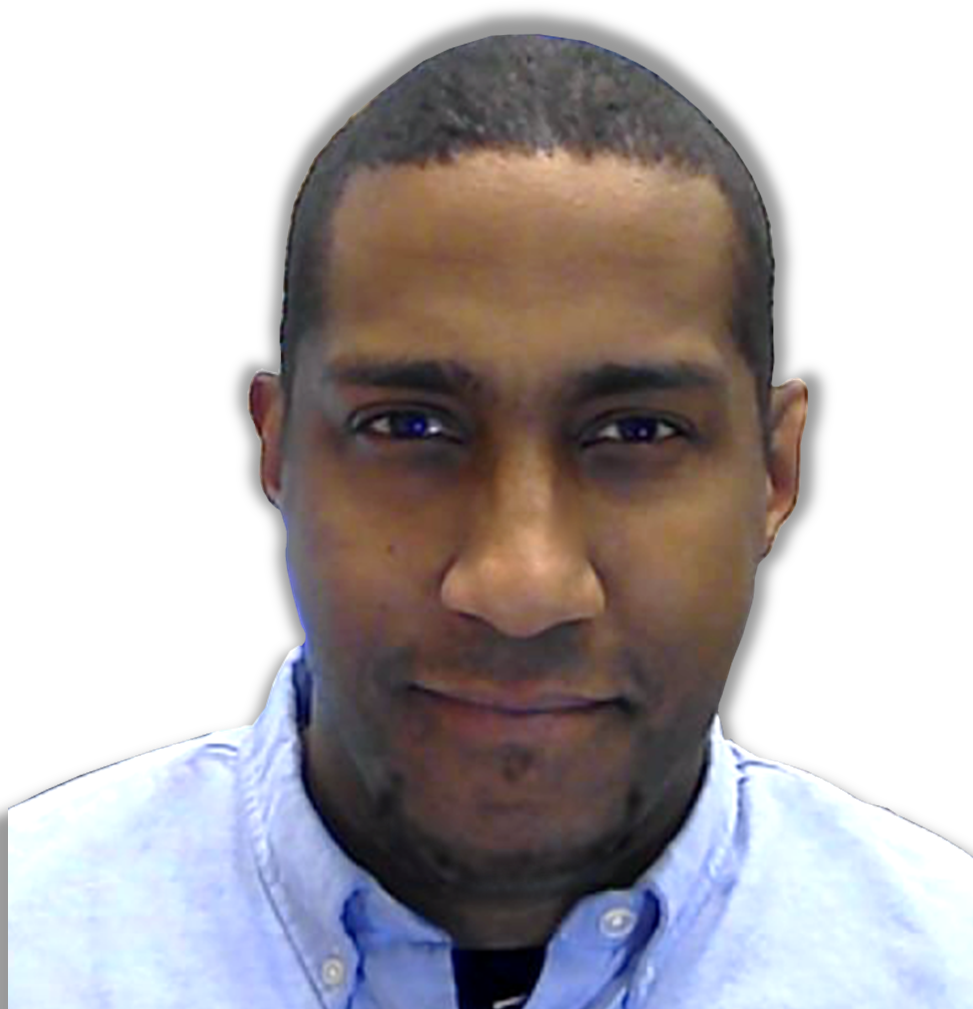}}]{Frank L. Hammond III}
Frank L. Hammond III (Member, IEEE) received a B.S. in electrical engineering from Drexel University in 2002, M.S degrees in electrical and mechanical engineering from the University of Pennsylvania in 2006, and a Ph.D. in mechanical engineering from Carnegie Mellon University in 2010. He worked as a postdoctoral research follow at Harvard University from 2010-2014 and a postdoctoral researcher at MIT from 2014-2015. He is currently an Assistant Professor of mechanical and biomedical engineering at the Georgia Institute of Technology. Dr. Hammond is a member American Society of Mechanical Engineering (ASME) and the Biomedical Engineering Society (EMBS). His research interests include the development of soft robotic actuators and sensors, fluidic circuits, wearable rehabiliation devices, teloperated maniupulators, and human augmentation.
\end{IEEEbiography}
\end{document}

%% file: introduction.tex
\begin{figure}[t!] 
\centering
\includegraphics[width=1.\linewidth]{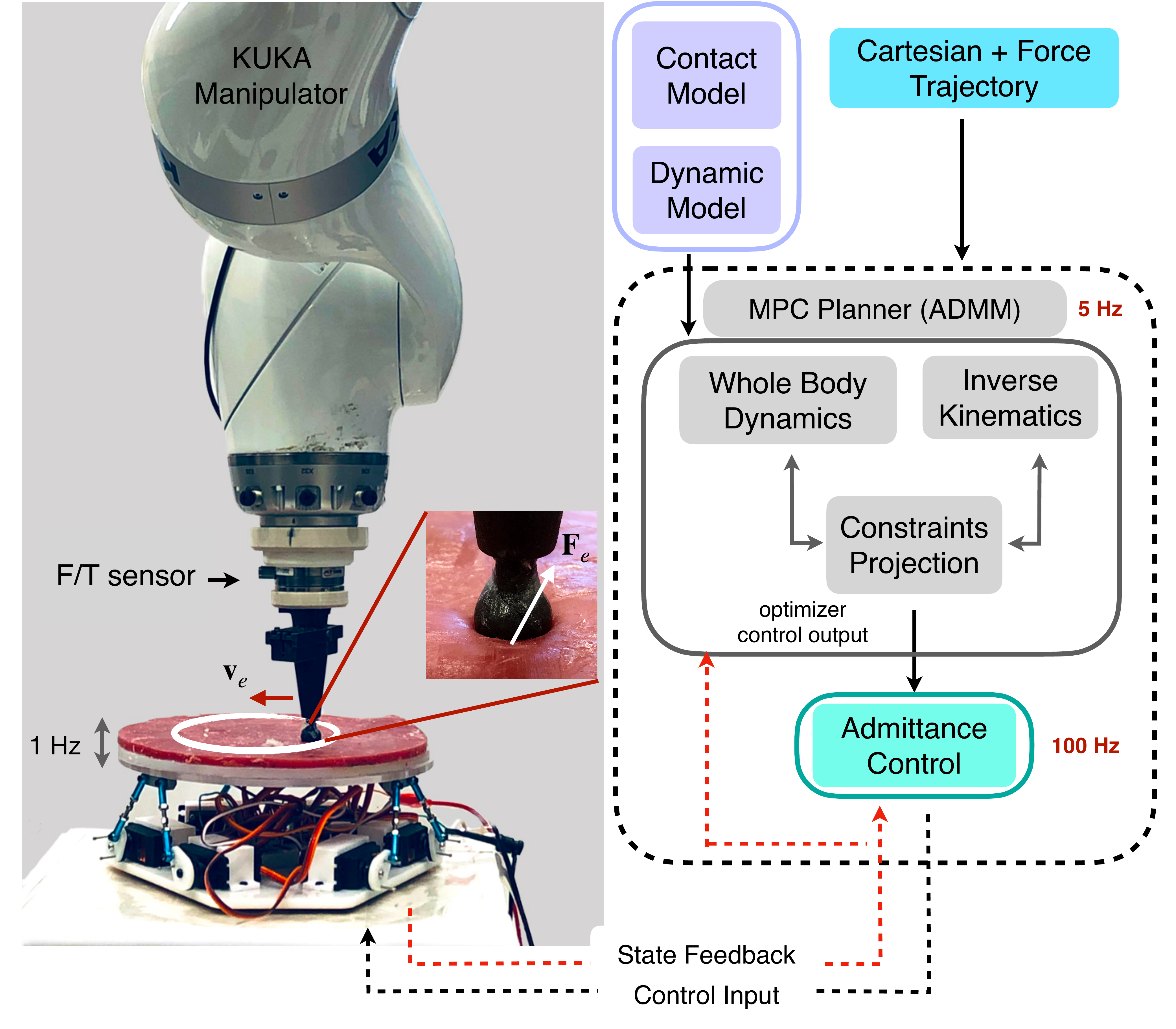}
\caption{KUKA manipulator experimentation setup. A manipulator performing a force-controlled motion task on a soft surface.}
\label{fig:setup_intro}
\end{figure}
\section{Introduction}

\IEEEPARstart{R}{obotics} applications in the medical domain have gained increasing attention over the past few decades \cite{Solis2016,Shademan2016}, where the planning and control of interaction forces between a robot and its environment are essential to various safety-critical tasks. For instance, interaction forces should be modulated accurately in compliant environments, such as surgical settings, micro-assembly, or biological tissue manipulation. Furthermore, force control based on identifiable physical models is crucial to identify instability modes (e.g., those caused by the bandwidth and system structure) and maintain reliable force interaction to guarantee safety. Thus, a model-based trajectory planning method with a high-fidelity contact model is essential for successful deployment with satisfactory motion and contact force tracking performance. 

\revisedtwo{In the field of model-based planning and simulation, there have been extensive studies in handling rigid contact and deformable contact models. For instance, rigid contact can be modeled using numerical solutions to complementarity problems \cite{posa2014direct, Pfrommer2020ContactNets:Representations, Anitescu2002AFriction, Stewart2000Rigid-bodyImpact}, or hybrid models with impact dynamics \cite{mirtich1996impulse, Featherstone2008RigidAlgorithms}.} Unlike in rigid body contact, soft contact models are subject to challenges posed by non-linear material properties and non-uniformity as well as intensive computation burden due to numerical computation for solutions. \revised{Considering soft contact mechanics during a physical interaction is important for safe adaptation and planning, especially in robotic surgical applications.} 
\revised{Soft contact modeling is challenging as it depends on material properties, deformation, and rolling friction.} Numerous contact models have been presented in the literature to model interactions involving elastic deformation\revised{ \cite{elandt2019pressure, Gilardi2002LiteratureModelling, Xu2019CompareStrategies, Mu2019RoboticMotion, Howe1996PracticalManipulation}}. These models have broad applications and are essential in many engineering areas such as machine design, robotics, multi-body analysis, to name a few. For contact problems that involve elasticity, Hertz adhesive contact theory has been well established\cite{Johnson1985ContactMechanics}. In this study, we focus on robotic tasks interacting with soft tissues, the contact behavior of which is determined by not only external and viscous forces, and contact geometry, but also material properties (see Figure \ref{fig:setup_intro}).

Many robotic tasks require motion planning in the presence of contact in constrained environments. \revised{For example, in the automation of medical applications, the environment is heavily constrained and safety is paramount \cite{Jarc2015}. In such scenarios, the robot motion and the reaction force imparted on the environment are coupled together. Similar examples are found in automated manufacturing, legged locomotion, and assistive robotics \cite{Park2019ActiveLearned, Suarez-Ruiz2018CanChair,Kemp2007ChallengesSettings}.}
As a promising approach along this direction, trajectory optimization (TO) with \revised{either hard or soft complementarity constraints \cite{posa2014direct,mordatch2012discovery,1631739,manchester2020variational,Patel2019Contact-implicitCollocation,sleiman2019contact,howell2022calipso} or explicit contact models \cite{neunert2017trajectory,neunert2018whole,8594284,8794250,kurtz2022contact}} has been extensively investigated in the robotics community. By incorporating the \revised{contact forces and joint states} into the optimization, contact-dynamics-consistent motions can be planned for complex robot behaviors, such as dynamic locomotion or dexterous object manipulation. \revised{While the above works demonstrated impressive results on automatically discovering contact sequences, the optimized contact force was not used for control, and only served as a way to explore the physical interaction with objects. In this work, we aim to accurately track a nominal force trajectory along with an end-effector path, which is commonly demanded but underexplored in safety-critical tasks.}

\revised{Safety-critical tasks such as soft material manipulation and medical applications require extensive manipulation with accurate contact interactions.} These include interaction with humans in proximity or with direct physical contact. \revised{In approaching the manipulation interaction problem, various methods have been explored, including model-based and model-free methods.} Data-driven techniques have been studied to learn the interaction between robotic manipulators and the environment \cite{Pereida2018Data-EfficientTransfer, ParigiPolverini2019MixedApproach,wang2020swingbot, han2021learning}. Unlike rigid contacts, soft environments are prone to both spatial and temporal uncertainties and it is challenging to learn the contact model and robot dynamics simultaneously through data \cite{Fazeli2017LearningImpact, ParmarFundamentalDynamics,Xu2019CompareStrategies}. 
\revised{In contrast, model-based approaches could capture the principle component of contact dynamics to a reasonable fidelity and be used along with an appropriate planning framework\cite{Dominici2014ModelSurgery, Kuppuswamy2020FastSensors, Long2014RoboticValidation}. 
These studies present control frameworks and sensor modalities that incorporate contact models to make use to make informed control decisions, including physics-based contact models\cite{Dominici2014ModelSurgery}, mesh-based simulation models, and tactile models\cite{Kuppuswamy2020FastSensors}. However, in these studies, simultaneous motion and force tracking is largely underexplored.
Our method models the contact with normal and frictional components of the force as a smooth dynamical system to augment the robot dynamics. Then, we embed it into a contact-aware TO for high-fidelity planning.} 


Since contact-rich environments are prone to disturbances and difficult to model or predict, \revised{it is indispensable to solve the aforementioned optimization problem online with updated contact state information. However, efficiently solving highly constrained TO with full robot and contact dynamics remains an open problem. Instead of solving a large-scale optimization in a holistic way, we propose a distributed optimization framework inspired by works from the legged locomotion community \cite{herzog2016structured, budhiraja2019dynamics,ZhouAcceleratedDynamics, zhou2022momentum, zhao2022adversarially}, but employ a different way for decomposing the original problem given the soft manipulation task. 
More specifically, this framework efficiently alternates between three sub-problems/sub-blocks: (i) an unconstrained sub-problem that only incorporates coupled rigid body (i.e., the robot end-effector) and soft contact dynamics for acurate force tracking; (ii) an inverse kinematics (IK) sub-problem that generates a joint trajectory to track the desired end-effector trajectory; (iii) a constraint projection sub-problem for handling inequality constraints including contact constraints. The proposed distributed framework enables an online execution in an model predictive control (MPC) fashion. Furthermore,} to maintain the contact and stability, an additional control layer will need to be deployed on top of the high-level trajectory planner. We add a low-level admittance force controller \cite{Wijayarathne2020IdentificationSurface} to handle uncertainites arising from the model as well as from disturbances in the environment.


The main contributions of this work are listed below:
\begin{itemize}
\item Presentation of a dynamic interaction model based on soft contact mechanics for a predefined geometry with Hertz visco-static theory.
\item Incorporation of the interaction model into a constrained TO to generate the desired cartesian path and force profile in an efficient, distributed fashion.
\item Experimental validation of the derived contact dynamical model and real-time implementation of the proposed TO algorithm with model predictive control (MPC).
\end{itemize}

A conference version of the work presented in this paper was published in \cite{wijayarathne2020simultaneous}. The work presented here extends the previous work in three respects. First, \revised{we extend the previous distributed framework into three sub-problems instead of only two, i.e., an independent IK sub-problem is extracted from the previous dynamics block to further split the dynamics (force tracking) and kinematics (end-effector tracking) planning.} 
\revised{We benchmark different versions of solving the same constrained TO problem including the one proposed in our conference version \cite{wijayarathne2020simultaneous}.} Second, our proposed trajectory planner is \revised{executed online} in an MPC fashion. We experimentally demonstrate the efficacy of this framework on simultaneous motion and force tracking tasks on a static and dynamic disturbance-induced platform. Last but not least, we implement a low-level controller to aid the high-level trajectory planner in motion and force trajectory tracking.

%% file: related_work.tex
\section{Related Work}

\label{sec:related work}
\textbf{Contact Models. }
Elastic contact mechanics \cite{Johnson1985ContactMechanics} have been extensively studied in various research fields where contact modeling is imperative for safety and performance requirements. Existing works in \cite{Gilardi2002LiteratureModelling, Schindeler2018OnlineLinearization,Lefebvre2003EstimatingMotion,Marhefka1999ASystems} have used soft contact models for both modelling and control. These works include quasi-static assumptions and studies of  \cite{Pappalardo2016Hunt-CrossleySurgery, Sun2018InternationalDamping} explore cases where high-velocity impacts on soft material are considered. In the impact cases, visco-elastic models have been widely investigated. For instance, studies in \cite{Pappalardo2016Hunt-CrossleySurgery,Gilardi2002LiteratureModelling} compared various visco-elastic models with experimental validations. A majority of these works show that the Hertzian-based Hunt-Crossey model is the one most suitable for visco-elastic cases.
Furthermore, fundamentals of frictional sliding motion are established in the works of  \cite{Howe1996PracticalManipulation, mason1986mechanics}, where the main focus is on rigid body contacts but generalizable to soft contacts. More recent works in \cite{Mu2019RoboticMotion, Fazeli2017ParameterContact} propose contact-area-based models.

\par

\textbf{Trajectory Optimization.} Trajectory optimization (TO) is a powerful tool to generate reliable and intelligent robot motions. Various numerical optimization methods have been proposed to solve a TO \cite{betts1998survey, jacobson1970differential, betts2010practical}. Among them, Differential Dynamic Programming (DDP) \cite{mayne1966second} and its first-order variant iterative Linear Quadratic Regulator (iLQR) \cite{li2004iterative} have aroused much attention in solving TO in the context of unconstrained problems, where only dynamics constraint is enforced. The Ricatti-like backward pass in DDP or iLQR effectively reduces the complexity by solving an approximated LQR problem over the entire horizon, and the optimization is solved in an iterative fashion. In \cite{tassa2012synthesis}, DDP is used in a balancing task of a humanoid robot with high degrees of freedom (DoFs). A follow-up work \cite{koenemann2015whole}, demonstrates a Model Predictive Control (MPC) implementation based on DDP. However, incorporating \revised{additional} constraints in standard DDP algorithms is still an open problem. In \cite{tassa2014control, Xie2017DifferentialConstraints,plancher2017constrained,7989016,howell2019altro,9561530,9134911,9340883,el2021equality,9560795,jallet2022constrained}, DDP-type variants are proposed to cope with equality or inequality constraints containing state or control variables. \revised{Meanwhile, an augmented Lagrangian (AL)-based method, Alternating Direction Methods of Multipliers (ADMM) \cite{boyd2011distributed}, is proposed to solve constrained optimization problem in a distributed fashion. In \cite{o2013splitting,o2016conic}, ADMM is used to solve large-scale convex problems with box and cone constraints. Although the convergence is only guaranteed for convex problem, ADMM has been proven effective for solving highly non-convex problems such as collision avoidance \cite{sindhwani2017sequential,zhao2021sydebo}, mixed-integer programming \cite{shirai2022simultaneous,lin2022multi}, linear complementarity problem \cite{aydinoglu2022real}, and recently employed for legged robot trajectory planning \cite{budhiraja2019dynamics,ZhouAcceleratedDynamics}. Our approach leverages both the distributed nature of ADMM and the efficiency of DDP, which decomposes the original problem into unconstrained whole-body dynamics planning with a soft contact model, kinematics planning, and constraint projection.}   


\textbf{Trajectory Optimization with Soft Contact. } 
Contact-aware TO \revised{for legged locomotion and dexterous object manipulation} is often built upon the assumption of rigid contact dynamics \cite{posa2014direct, mordatch2012discovery, 1631739, manchester2020variational, Patel2019Contact-implicitCollocation, sleiman2019contact, neunert2017trajectory,neunert2018whole,8594284,8794250,kurtz2022contact}. In \cite{marcucci2016two, neunert2017trajectory, neunert2018whole}, soft contact models are integrated into the system dynamics to approximate the hard contact, and the contact models are still relatively simple. In \cite{fahmi2020stance}, a soft contact model was considered in the optimization formulation for whole-body locomotion control. \revised{Although incorporating these soft contact models has demonstrated impressive results, it is challenging to directly use these approaches for soft manipulation tasks considered in this work. Because most of them assume simple spring-damper type soft contact models, which still largely mismatch the contact surface deformation or elasticity \cite{gilardi2002literature, Dominici2014ModelSurgery, Kuppuswamy2020FastSensors, Long2014RoboticValidation} in reality.} 
Therefore, advanced planning algorithms that accurately model complex contact dynamics are imperative to enable maneuvering over complex terrain or grasping irregular objects. To date, \revised{constrained} TO incorporating a high-fidelity deformable contact model remains underexplored in the field.


\par

\textbf{Model Predictive Control (MPC). }
The generation of trajectories for a given model and fixed horizon is computationally prohibitive in nature, making it difficult to deploy in real-time contact-rich applications, where model uncertainties and environmental disturbances are  ubiquitous. Model Predictive Control (MPC) is a powerful strategy widely used to generate motion plans in real time and be adaptive to state changes due to environmental disturbances \cite{farshidian2017real, Cortesao2009MotionFeedbacke}. Recent advances in fast automatic differentiation (AutoDiff) \cite{Frigerio2016RobCoGenLanguages} and AutoDiff compatible rigid body models has enabled real time optimal control. The study in \cite{Kleff2020High-FrequencyManipulator} showed a hardware implementation of MPC with a DDP optimizer framework at an update rate of $1000$ Hz on a 7-DOF robot for a vision based point-to-point trajectory planning. 

\textbf{Admittance Control. }
To cope with un-modeled modalities of the contact, we use a low-level force controller \revised{(FC)} based on admittance control\cite{Wijayarathne2020IdentificationSurface} which has a fast control update rate compared to the high-level planner. It can compensate for the uncertainties that arise spatially over the surface (e.g., stiffness, slipperiness, and damping). Admittance control has been long studied and proven to work efficiently in compliant environments\cite{Ott2010UnifiedControl}. Furthermore, low-level controller mitigates instabilities\cite{10.1115/1.2897725} arises from the contact caused by the control update rate, stiffness mismatch, and high gains.

\par

%% file: methods_contact.tex
\section{Deformable contact modeling}
\subsection{Contact modeling via Hertz's theory}
\label{contact interaction model}
\begin{figure}[t]
\centering
\includegraphics[width=0.8\linewidth]{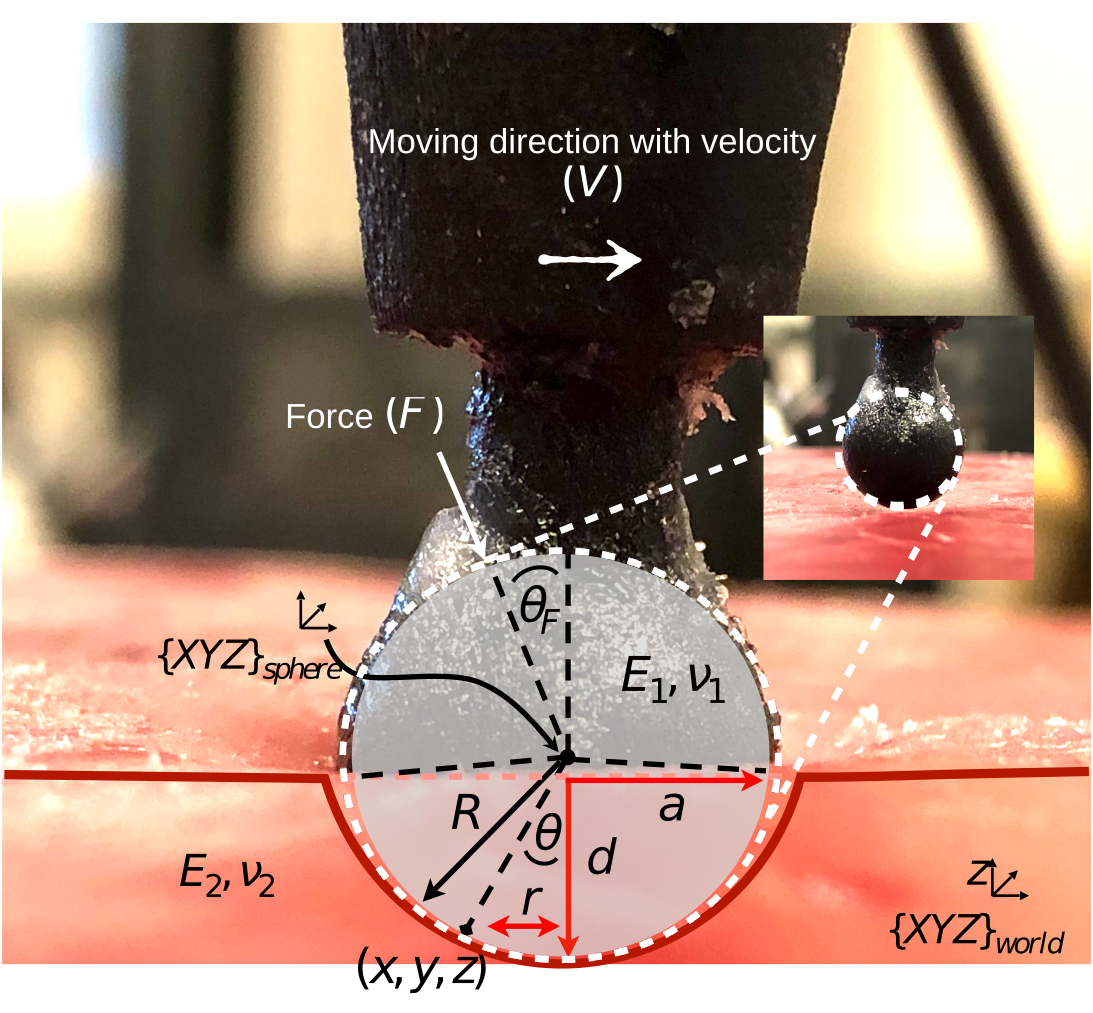}
\caption{A graphical illustration of the soft contact model between surface and the end-effector tool}
\label{fig:contact_ball}
\end{figure}

In this section, we model the interaction dynamics between an application tool mounted on a manipulator and a soft tissue in terms of contact geometry and mechanics. \revised{For simplicity, the tool tip of this study is of a
spherical} \revised{shape}  (but not limited to). \revised{We} assume that the tool \revised{end tip} used is rigid and \revised{relatively stiffer} compared to the contact surface. With these assumptions, we derive a \revised{contact} dynamical model based on the contact friction theory and pressure distribution. According to Hertz's theory, the largest static \revised{deformation} is \revised{observed} at the \revised{mid} point of the circle (\revised{as shown in the deformation in} Figure~\ref{fig:contact_ball}) and can be expressed as:
 \begin{equation}
     d=\left[\frac{9F^{2}}{16E^{2}R}\right]^{\frac{1}{3}}
     \label{eq:quasi_static_model}
 \end{equation}
where $E$ is the reduced Young's modulus of the tool and surface, $R$ is the radius of the tool end, and $F$ is the force imparted on the surface by manipulator \revised{tool-tip}. Combined Young's modulus of the \revised{tool-tip} and the soft contact surface material can be lumped to one term as:
 \begin{equation*}
     \frac{1}{E}=\frac{1-\nu_1^2}{E_1}+\frac{1-\nu_2^2}{E_2}
 \end{equation*}
where $E_1$, $E_2$, and $\nu_1$, $\nu_2$ are Young's moduli and Poisson ratios of the end-effector and contact surface material, respectively.
In our scenario, we assume the contact part as a rigid object, and thus Young's modulus of the spherical \revised{tool-tip} $E_1$ is \revised{comparatively high}. \revised{Consequently, the lumped stiffness can be approximated as}  \revised{$E=E_2/(1-\nu_2^2)$}.

The deformation and stress distributions on the \revised{contact} surface are approximated by the universal Hooke's law and Hertz's theory. \revised{A detailed elaboration} of normal, radical, and hoop (i.e., moving direction) stress distributions within the contact area in the cylindrical coordinate system are provided in the Appendix \revised{\ref{appendix:deformable_contact}}.



The deformation distribution is derived from the stress distribution equations as follows:
\begin{equation*}
\sigma_n=\begin{cases}
  \frac{3\pi}{8a}\Big[\frac{1-\nu^2}{E}\Big]p_m(2a^2-r^2), & (r\leq a) \\ \\
  \frac{3}{4a}\Big[\frac{1-\nu^2}{E}\Big]p_m\Big[(2a^2-r^2)\sin^{-1}\big(\frac{a}{r}\big) \\ \quad + a(r^2-a^2)^{\frac{1}{2}}\Big], & (r \geq a)
    \end{cases}
\end{equation*}
where $p_m=F/(\pi a^2)$ is the average stress applied in the contact part by the tool-tip and $a=\sqrt{Rd}$ is the radius of contact area (see Figure~\ref{fig:contact_ball}). \revised{A force vector $F$ at an angle $\theta_F$ to the normal is applied to the application tool and moves in a curved path of a radius $R$ with a uniform velocity $\mathbf{v}_e$ in the frame $\{\text{sphere}\}$}
. \revised{This represents a scenario of a tool interaction with a soft surface to accomplish a motion task}. For \revised{the sake of simplicity}, our model focuses on sliding friction (primary mode) and ignores other frictional sources such as adhesion and rolling induced by deformation. \revised{This assumption is valid for modeling purposes where adhesion and rolling are secondary and specific to the tool material and the application.}
 \par \revised{To derive frictional forces, we use principle stress on the contact.}
\revised{If $\sigma_\theta$ represents the principal stress within the contact circle due to the symmetry of our contact scenario, we can represent the stress tensor of any contact point $(r,\theta,z)$ in cylindrical coordinates relative to frame $\{\text{sphere}\}$ via the Cauchy stress theory \cite{Johnson1985ContactMechanics}.}
\begin{equation}
  \vg \sigma= {\left[ \begin{array}{ccc}
\sigma_r & 0 & \sigma_{rz}\\
0 & \sigma_\theta & 0\\
\sigma_{zr} & 0 & \sigma_z
\end{array} 
\right ]}
\end{equation}
Since the task is defined in the Cartesian frame, we convert parameters to Cartesian coordinates from cylindrical coordinates. The stress tensor in Cartesian coordinate is $\vg \sigma_c= T^T \vg \sigma T$, where the transformation matrix $T$ is defined in Appendix \ref{appendix:deformable_contact}.

At an arbitrary point on contact surface $(x, y, z)_{\{\text{sphere}\}}$, the normal vector from this point to centroid of spherical cap is $\mathbf{n}=[s\theta \quad 0 \quad c\theta]^T$.\footnote{we denote $\cos \theta = c \theta$ and $\sin \theta = s\theta$.} Then, the normal stress of the contact surface is $\sigma_n=\mathbf{n}^T\vg \sigma_c \mathbf{n}$ with
\begin{equation*}
\sigma_n = \sigma_r c^2\theta s^2\theta + \sigma_\theta s^4\theta+ \sigma_z c^2\theta+2 \sigma_z s\theta c^2\theta
\end{equation*}
Given this stress expression, the overall friction force of the contact surface is represented as
\begin{align}
    df=\mu \sigma_n dS=\mu \sigma_n\times 2\pi r \frac{dr}{c\theta} \\
    f=\int dfc\theta= 2\pi \mu \int_{0}^{a} \sigma_n rdr
    \label{eq. friction_integral}
\end{align}
where, $df, dr, dS $ are the differential elements of the friction, $r$ and contact area. In the surface normal direction, it is assumed that the surface is in contact with the end point of the tool.  As a result, Eq. ~(\ref{eq:quasi_static_model}) always holds. \revised{To derive the dynamic model in the normal direction of contact}, the derivative form of Eq.~ (\ref{eq:quasi_static_model}) is taken
\begin{equation}
    \dot{z}=-\dot{d}=-\left[\frac{1}{6E^{2}RF_z}\right]^{\frac{1}{3}}\dot{F_z}
    \label{eq:quasi-static-dir}
\end{equation}
where $z$ represents the position along the surface normal direction of the contact point and force along the normal direction is defined as $F_z = F\cos{\theta_F}$.
In the moving direction, Eq. ~(\ref{eq. friction_integral}) and $\mu F_z$ give the frictional force caused by the normal force $F_z$, which is:
\begin{equation}
    \v F_f=f=\mu F_z\left[1+(2\nu-1)\frac{3a^2}{10R^2}\right] \v n_v  + k_d\v v_e
    \label{eq:friction_model}
\end{equation}
where $k_d$ is a damping coefficient in the moving direction. By substituting $a=\sqrt{Rd}$ and Eq.~(\ref{eq:friction_model}), we have the derivative form of Eq.~(\ref{eq:friction_model}). $\v n_v$ is the unit vector of the velocity and $\v v_e$ is the end-effector velocity at the contact point.

\begin{equation}
    \label{eq:friction_dyn}
    \dot{\v F}_f=\dot{f}=\mu \dot{F_z}+\frac{3 \mu (2\nu-1)}{10R}\left(\dot{F_z}d-F_z\dot{z}\right)\v n_v + k_d\dot{\v v}_e
\end{equation}

The overall model, \revised{in simplistic terms}, \revised{$ \dot{\mathbf{F}}_e = \revisedtwo{\mathcal{G}}(F_z, \v v_e, \text{contact parameters})$} with the frictional and normal force components of the contact model can be written in a compact form as:

\begin{multline}\label{contact model}
    \dot{\mathbf{F}}_e = \Big(\left(6E^{2}RF_z\right)^{\frac{1}{3}}\dot{d}\Big)\mathbf{n}_z \\ + \Big(\mu \dot{F_z}+\frac{3 \mu (2\nu-1)}{10R}\left(\dot{F_z}d
    +F_z\dot{d}\right)\Big)\mathbf{n}_v + k_d\dot{\v v}_e
\end{multline}

 where $\mathbf{n}_v$ is unit vector and $d$ is the deformation at central point of contact circle and is calculated from Eq.~(\ref{eq:quasi_static_model}), and $\dot{d} = \dot{x}_z$. $F_z$ is the vertical force (the surface normal direction) applied on the surface by the manipulator and $\v v_e = || \v J \dot{\v q}||$  is the moving velocity of the tool contact point.

\begin{figure*}[t] 
\centering
\includegraphics[width=1\linewidth]{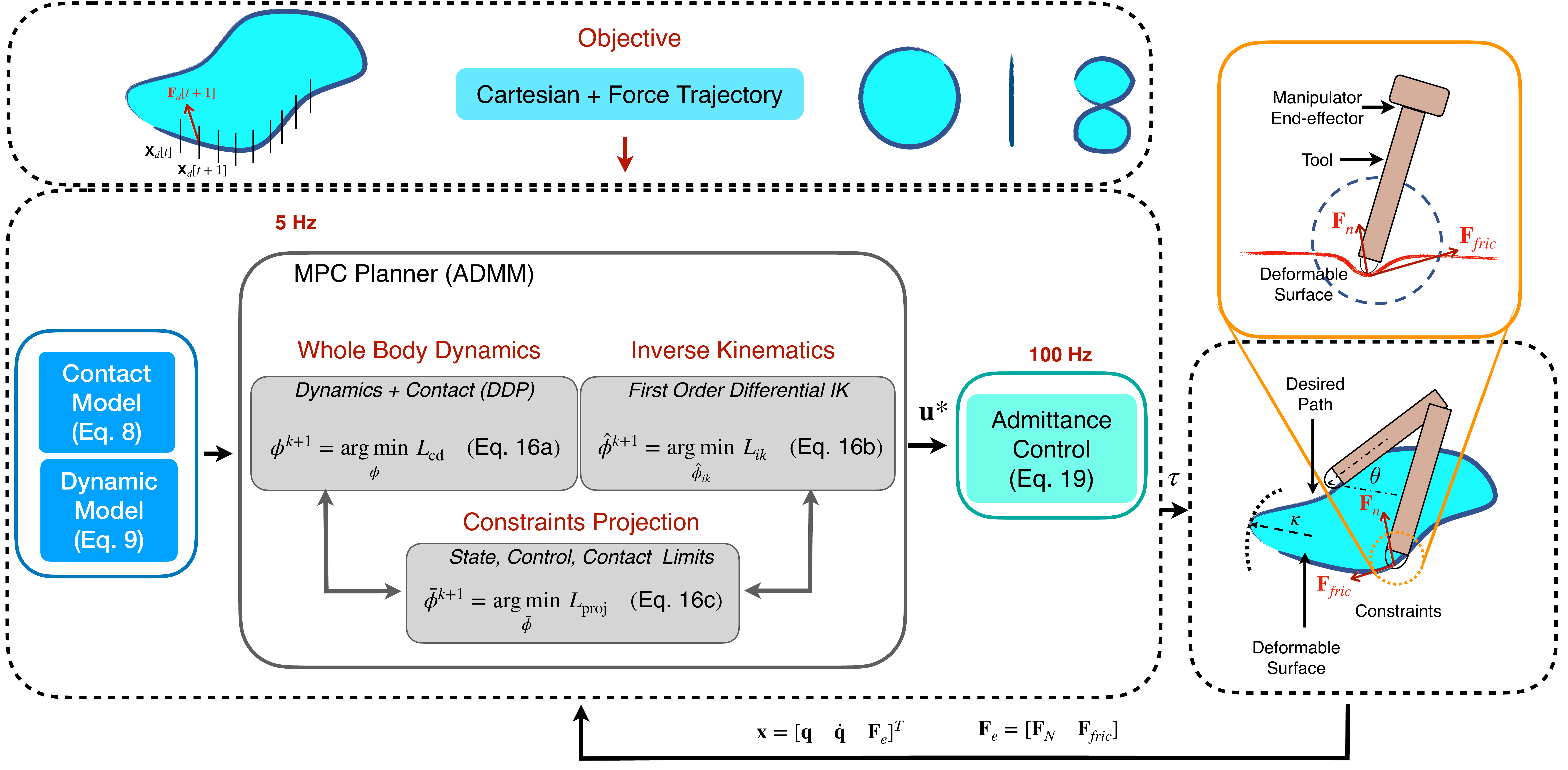}
\caption{Planning Framework. The planner is composed of an ADMM based solver executed in MPC fashion at $5$ Hz. ADMM solver contains three sub-blocks for optimization. Manipulator and contact dynamic models are used in the whole body dynamics block. Inverse kinematics block uses first-order differential kinematics to generate Cartesian trajectory. States are projected to satisfy the state and contact constraints in the projection block. The low-level force controller which is the admittance controller is executed at $100$ Hz. The desired Cartesian and force trajectory is tracked while obeying the constraints provided.}
\label{fig:overall_overview}
\end{figure*}

\begin{figure}[h] 
\centering
\includegraphics[width=1.0\linewidth]{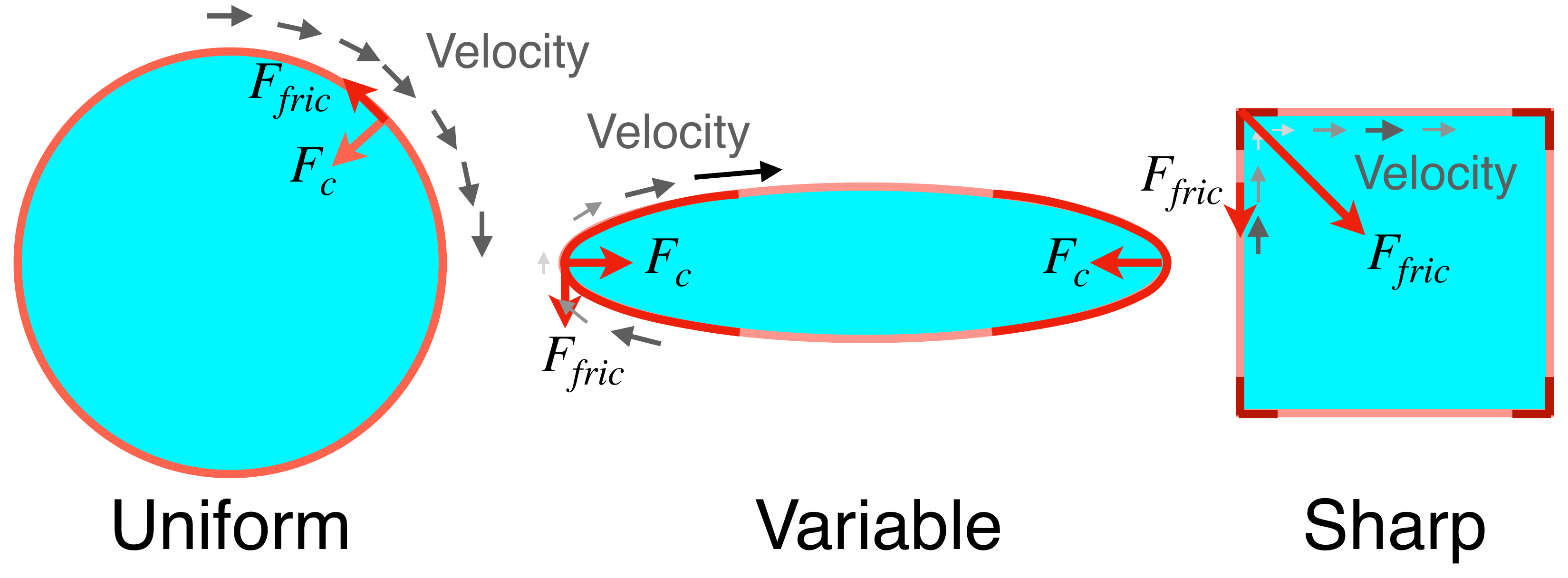}
\caption{Different Cartesian Path Geometries. At sharp corners of a Cartesian path, the robot needs to vary the velocity to compensate for centripetal and frictional force components caused through the contact}
\label{fig:path_curvatures}
\end{figure}

\revised{\subsection{Manipulator Dynamics}
The manipulator model dynamics is expressed below:
\begin{equation}
     \ddot{\v q} = \mathbf{M(\mathbf{q})}^{-1} (\boldsymbol{\tau}_u-\mathbf{C}(\mathbf{q},\mathbf{\dot{q}})\mathbf{\dot{q}} - \mathbf{G}(\mathbf{q})- \mathbf{J}^T\mathbf{F}_e)
    \label{eq:manip_model}
\end{equation}
where $\v q, \dot{\v q}, \ddot{\v q}, \vg \tau_u \in \R^k$ are the manipulator joint position, velocity, acceleration, and torque vectors. We use $k=7$ for our 7-DoF robotic arm. $\mathbf{M(\mathbf{q})}$ is the joint space mass matrix, $\mathbf{C}(\mathbf{q},\mathbf{\dot{q}})$ is the Coriolis term, $\mathbf{G}(\mathbf{q})$ is the gravity term, $\boldsymbol{\tau}_u$ is the torque applied at joints, $\mathbf{J}$ is the Jacobian with respect to the contact point, and $\mathbf{F}_e$ is the external Cartesian wrench at the end-effector.
}

\subsection{Contact Constraint Modeling}\label{subsec:contact_constraint_modeling}
The motion studied in this paper is primarily in the sliding mode, which an equality constraint can describe. Since the sliding is embedded in the contact model, additional constraints for sliding are not required. However, a constraint is added to make sure the robot is only sliding in the desired path where the path is curved. Figure~\ref{fig:path_curvatures} shows the two components of the force acting on the tool, namely, frictional and centripetal forces along the path of curvature $R_c$ as shown in Figure~\ref{fig:overall_overview}. The contact model provides the sliding friction, and the centripetal force constraint is added as a constraint:
\begin{align}
    \frac{\mathbf{J}^{-T}\mathbf{M}(\mathbf{q})\mathbf{J}^{-1} \| \v J \revised{\dot{\v q}}\|^2}{\kappa} &\le \mu \mathbf{N}^T\mathbf{F}_e\mathbf{N} 
    \label{eq:contact_constraints}
\end{align}
where $\v F_e \in \R^3$ is the force vector at the end effector and $\v J$ is the Jacobian.
$\kappa$ is the curvature of the motion path (as shown in Figure \ref{fig:overall_overview}), $\v J^{-T}\v M(\v q)\v J^{-1}$ is the effective mass at the contact point of the robot with a mass matrix of $\v M(\v q)$ and
$\v J\revised{\dot{\v q}}$ is the moving velocity of the contact point. $\v N$ is the surface average normal vector (as precept by the force-torque sensor). In this study, we use $\v N = \begin{bmatrix}0 & 0 & 1 \end{bmatrix}^T$, which is only in the $z$ direction\footnote{$\mathbf{N}$ varies with the surface deformation which is stochastic and can estimated through an external force-torque sensor attached to the end-effector}.
The constraint represented by Eq. \ref{eq:contact_constraints} keeps the robot in contact when the cartesian tracking path has a curvature, and the robot is operated in a lower impedance mode. It is an effect of the resulting centripetal force on the effective mass at the contact. For instance, if the velocity at a curve is high, it would slip in the orthogonal direction of the moving direction unless high positional gains are used to compensate for it.

%% file: methods_admm.tex
\section{Problem formulation}
\label{sec:overall dynamic model}
The optimization problem is to solve a control trajectory that would result in a desired cartesian trajectory along a desired force profile. \revised{The overall problem is formulated in Formulation \ref{formulation_1}, where the state is represented as $\mathbf{x} = [\v q \hspace{4pt} \dot{\v q} \hspace{4pt} \mathbf{F}_e]^T$ including the manipulator joint position, velocity, and the end-effector force vectors. The control input is equivalent to the joint torque vector $\mathbf{u} = \boldsymbol{\tau}_u$. For simplicity, we use $\vg \phi=(\v x[0,\ldots,N],\v u[0,\ldots,N-1])$ to represent the sequence of state-control pairs.} 

\revised{The objective function is comprised of a force tracking cost, an end-effector pose tracking cost, and a regularization term on the applied torques.} The normal force error with respect to \revised{the force} reference \revised{$F^d$} is expressed as $\delta\mathbf{F}[i] = ( F_z[i] - \revised{F^d[i]})$. Matrices $\mathbf{Q} \in \mathbb{R}^{n\times n}$ and $\mathbf{R} \in \mathbb{R}^{m\times m}$ are the state and control weighting matrices, $\mathcal{FK} \in \mathbb{R}^{4 \times 4}$ is the forward kinematics function of the manipulator, and $W_p \in \mathbb{R}^{4\times 4}$ is the state weighting matrix for the \revised{pose tracking cost with respect to a desired end-effector trajectory $\v x_{e}^d$}.  
\begin{formulation}
\caption{Simultaneous trajectory and force optimization}\label{formulation_1}
\begin{subequations}\label{eq:prime_objective}
\begin{align}\label{eq:local_cost}
    \nonumber &(\textit{\rm{Tracking Task}}) \quad \underset{\vg \phi}{\text{min}} \ \sum_{i = 0}^{N} \hspace{2pt}  \overbrace{\delta\mathbf{F}[i]^T\mathbf{Q}_F\hspace{2pt}\delta\mathbf{F}[i]}^{\text{force tracking}}
    + \mathbf{u}[i]^T\mathbf{R}\mathbf{u}[i]  \\ \nonumber 
    &\hspace{3.5cm} + \overbrace{W_p\big\| \mathcal{FK}(\revised{\v q}[i])-\mathbf{x}_e^d[i]\big\|_2^2}^{\text{pose tracking}} \\ \nonumber
    &(\textit{\rm{Decision Variables}}) \hspace{0.5cm} \vg \phi[i]=[[\overbrace{\revised{\v q}[i] \hspace{4pt} \dot{\revised{\v q}}[i] \hspace{4pt} \mathbf{F}_e[i]}^{\v x[i]}]^T, \v u[i]^{T}]^T \\ 
    &\hspace{3.5cm} \forall i = 1, \ldots, N-1 \\
    \label{eq:system_dynamics}
    &(\textit{\rm{Dynamics}}) \hspace{1cm} \text{s.t.} \quad 
    \mathbf{x}[i+1] = \revised{\v f}(\mathbf{x}[i], \mathbf{u}[i])\\
    \label{eq:init_condition}&(\textit{\rm{Initial Condition}}) \hspace{0.85cm}
    \mathbf{x}[0] = \mathbf{x}_0 \\\label{eq:joint_limit}
    &(\textit{\rm{Joint Limits}}) \hspace{1.5cm} \revised{\v q_{\text{lower}}} \leq \revised{\v q}[i] \leq \revised{\v q_{\text{upper}}} \\\label{eq:wrench_limit}
    &(\textit{\rm{Torque Limits}}) \hspace{1.2cm} \revised{\v u_{\text{lower}}} \leq \v u[i] \leq \revised{\v u_{\text{upper}}}  
    \\\label{eq:sliding_limit}
    &(\text{Contact Constraint})
    \hspace{0.5cm}
    \frac{\mathbf{J}^{-T}\mathbf{M}(\mathbf{q})\mathbf{J}^{-1} \| \v J \revised{\dot{\v q}}\|^2}{\kappa} \le \mu \mathbf{N}^T\mathbf{F}_e\mathbf{N}
\end{align}
\end{subequations}
\end{formulation}


\revised{After combining the contact dynamics model in Eq.~(\ref{contact model}) and the manipulator dynamics model in Eq.~(\ref{eq:manip_model}), the overall system dynamics can be discretized through the Euler method and written as function $\v f(\cdot,\cdot)$ in Eq.~(\ref{eq:system_dynamics}). The initial condition $\v x_0$ is given as shown in Eq.~(\ref{eq:init_condition}). The joint and torque limits are defined in Eq.~(\ref{eq:joint_limit}) and Eq.~(\ref{eq:wrench_limit}) with lower and upper bound pairs ($\v q_{\text{lower}}$,$\v q_{\text{upper}}$) and ($\v u_{\text{lower}}$,$\v u_{\text{upper}}$), respectively. The sliding contact constraint in Eq.~(\ref{eq:sliding_limit}), discussed in Sec.~\ref{subsec:contact_constraint_modeling}, also needs to be incorporated into the optimization problem. Next, we describe how we solve Formulation \ref{formulation_1} in a distributed manner, i.e., the proposed ADMM algorithm.}

\section{Constrained Trajectory Optimization with Contact Dynamics}
\label{sec:admm_opt}

Given the manipulator and the contact dynamics models, 
Differential Dynamic Programming (DDP) is used to generate desired joint and Cartesian motion as well as force profiles obeying \revised{the system dynamics defined by $\v f(\cdot, \cdot)$}. DDP is well received for effectively solving unconstrained trajectory optimization \cite{jacobson1970differential}. It represents an indirect method which only optimizes control inputs, and the dynamics constraint is implicitly satisfied during the forward trajectory rollout. Given an initial guess of control inputs, an updated state trajectory is generated by forward propagating the differential equation of system dynamics. Then a quadratic approximation is constructed for the cost function and dynamics around the current trajectory, so that a Riccati recursion can be used to derive the optimal feedback control law. By iteratively updating the state and control trajectories, the optimization will converge to an optimal solution. 

One limitation of DDP stems from its difficulty in addressing constraints other than the dynamics constraint enforced during the forward pass roll-out. Since our contact model enforces state, control, and frictional constraints, it is desired to incorporate these contact constraints along with the state and control constraints. Our previous work in \cite{wijayarathne2020simultaneous} proposed an iterative and distributed method based on Alternating Direction Method of Multipliers (ADMM) to incorporate the contact dynamics and \revised{additional} constraints. In this work, we introduce an inverse kinematics sub-problem and extend the entire ADMM framework to be a consensus variant to further improve the computational efficiency. Note that a sequential variant can also be established. Details are demonstrated in \revised{Appendix \ref{appendix:admm}} and we benchmark multiple variants in \revised{Sec.~\ref{subsec:TO_benchmark}}.

The ADMM algorithm decomposes a large-scale, holistic optimization problem into sub-problems and solves each sub-problem iteratively. In each iteration, \revised{the primal and dual problems} are solved sequentially. Under mild conditions, \revised{all variable sets coming from primal and dual problems} converge to the optimal solutions. More details about ADMM algorithm are referred to \cite{boyd2011distributed}. To apply this algorithm for our soft-contact manipulation problem, we define various sets of copies and the corresponding consistency constraints. Therefore, the original Formulation \ref{formulation_1} is transcribed into a distributed version as shown in \revised{Formulation \ref{formulation_admm_consensus}}. For simplicity, we define $\vg \lambda=({\revised{\dot{\v q}}}^T,{\v F_e}^T)^T$ and $\bar{\vg \phi}=(\revised{\bar{\v q}}[0,\ldots,N],\bar{\v u}[0,\ldots,N-1],\bar{\vg \lambda}[0,\ldots,N])$\footnote{The decision variables $\revised{\v q}$ and $\vg \lambda$ are subsets of the full state $\v x$} to express the concatenated copies of states and controls that are required to be projected. Meanwhile, the variable set $\hat{\vg \phi}=(\revised{\hat{\v q}}[0,\ldots,N])$ denotes a copy of joint position $\v x_M$ and handles the end-effector tracking, i.e., the inverse kinematics (IK). The closed and convex sets $\mathcal{J}$, $\mathcal{U}$ and $\mathcal{F}$ stand for joint limit (\ref{eq:joint_limit}), control limit (\ref{eq:wrench_limit}) and contact constraint (\ref{eq:sliding_limit}), respectively. By utilizing an indicator function \revised{which specifies a zero value if a targeting variable is within a feasible set and an infinitely large value if not}, the above constraints are encoded as a cost term $I_{\mathcal{J},\mathcal{U},\mathcal{F}}$ inside the new cost function in Formulation \revised{\ref{formulation_admm_consensus}}. 

\begin{formulation}[t]
\caption{Distributed constrained optimization (consensus)}\label{formulation_admm_consensus}
\begin{subequations}\label{eq:OCP_three_block}
\begin{align}\nonumber
    &(\textit{\rm{Tracking Task}}) \quad \underset{\vg \phi, \hat{\vg \phi},\bar{\vg \phi}}{\text{min}} \ \sum_{i = 0}^{N} \hspace{2pt}  \delta\mathbf{F}[i]^T\mathbf{Q}_F\hspace{2pt}\delta\mathbf{F}[i]
    + \mathbf{u}[i]^T\mathbf{R}\mathbf{u}[i]  \\ \nonumber 
    &\hspace{3.5cm} + W_p\big\| \mathcal{FK}(\revised{\v q}[i])-\mathbf{x}_e^d[i]\big\|_2^2 \\
    \nonumber &\hspace{3.5cm} +  I_{\mathcal{J},\mathcal{U},\mathcal{F}}(\revised{\Bar{\v q}}[i], \bar{\v u}[i],\Bar{\vg \lambda}[i]) \\ \nonumber
    &(\textit{\rm{Variables-DDP}}) \hspace{0.5cm} \vg \phi[i]=[[\overbrace{\revised{\v q}[i] \hspace{4pt} \revised{\dot{\v q}}[i] \hspace{4pt} \mathbf{F}_e[i]}^{\v x[i]}]^T, \v u[i]^{T}]^T \\ \nonumber
    &(\textit{\rm{Variables-IK}}) \hspace{0.84cm} \hat{\vg \phi}[i]=\revised{\hat{\v q}}[i]^T\\ \nonumber
    &(\textit{\rm{Variables-Proj}}) \hspace{0.6cm} \bar{\vg \phi}[i]=[\revised{\bar{\v q}}[i]^T, \bar{\v u}[i]^T, \bar{\vg \lambda}[i]^T]^T\\
    &\hspace{2.95cm} \forall i = 0, 1, \ldots, N-1 \\\label{eq:dynamics_constraint_consensus}
    &(\textit{\rm{Dynamics}}) \hspace{1.57cm} \text{s.t.} \quad 
    \mathbf{x}[i+1] = \revised{\v f}(\mathbf{x}[i], \mathbf{u}[i])\\
    &(\textit{\rm{Initial Condition}}) \hspace{1.4cm}
    \mathbf{x}[0] = \mathbf{x}_0 \\\label{eq:consistency_constraints_consensus}
    &(\textit{\rm{Consistency Constraints}}) \hspace{0.2cm} \begin{cases}\revised{\hat{\v q}} = \revised{\bar{\v q}}\\
    \revised{\v q} = \revised{\bar{\v q}} \\
    \v u=\bar{\v u} \\
    \vg \lambda = \bar{\vg \lambda}
    \end{cases}
\end{align}
\end{subequations}
\end{formulation}



Given the transcribed optimization problem, \revised{an augmented Lagrangian (AL) in a scaled form (See \cite{boyd2011distributed} Sec.~3.1.1) can be derived 
and written as follows:
\begin{equation}\label{eq:AL_three_block_consensus}
\begin{aligned}
    \mathcal{L}(\vg \phi, \hat{\vg \phi}, \bar{\vg \phi}, \v v) = &\sum_{i = 0}^{N} \hspace{2pt} (\delta\mathbf{F}[i]^T\mathbf{Q}_F\hspace{2pt}\delta\mathbf{F}[i]
    + \mathbf{u}[i]^T\mathbf{R}\mathbf{u}[i]) \\ & + \hspace{2pt} W_p\big\| \mathcal{FK}(\revised{\hat{\v q}}[i])-\mathbf{x}_e^d[i]\big\|_2 \\ \nonumber & +  I_{\mathcal{J},\mathcal{U},\mathcal{F}}(\revised{\Bar{\v q}}[i], \bar{\v u}[i],\Bar{\vg \lambda}[i])\\
    \nonumber & + \frac{\rho_j}{2}\|\revised{\hat{\v q}} - \revised{\bar{\v q}}+\v v_{\rm ik}\|_2^2 + \frac{\rho_j}{2}\|\revised{\v q} - \revised{\bar{\v q}}+\v v_j\|_2^2\\
    \nonumber &+ \frac{\rho_u}{2}\|\v u-\bar{\v u}+\v v_u\|_2^2 + \frac{\rho_f}{2}\|\vg \lambda - \bar{\vg \lambda}+\v v_f\|_2^2
\end{aligned}
\end{equation}}
\revised{where $\vg \phi$, $\hat{\vg \phi}$ and $\bar{\vg \phi}$ are primal variables. $\v v_{\rm ik}$, $\v v_j$, $\v v_u$ and $\v v_f$ are dual variables related to each consistency constraints defined in (\ref{eq:consistency_constraints_consensus}). $\rho_j$, $\rho_u$, $\rho_f$ are step-size parameters corresponding to each constraint. Note that since the first two consistency constraints possess the same projection goal $\bar{\v x}_M$, parameter $\rho_j$ is adopted for both constraints.}

Then based on ADMM, the original problem is divided into three sub-problems which are also known as sub-blocks. Each sub-block only requires part of the aforementioned AL as the local cost function:

\revised{Whole-body} dynamics sub-block:
\begin{equation}
\begin{aligned}
    \label{eq:block_1}
    \mathcal{L}_{\revised{\rm wbd}}\revised{(\vg \phi, \bar{\vg \phi}, \v v)} = &\sum_{i = 0}^{N} \hspace{2pt} (\delta\mathbf{F}[i]^T\mathbf{Q}_F\hspace{2pt}\delta\mathbf{F}[i]
    + \mathbf{u}[i]^T\mathbf{R}\mathbf{u}[i]) \\
    & + \frac{\rho_j}{2}\|\revised{\v q} - \revised{\bar{\v q}}+\v v_j\|_2^2+ \frac{\rho_u}{2}\|\v u-\bar{\v u}+\v v_u\|_2^2\\
    & + \frac{\rho_f}{2}\|\vg \lambda - \bar{\vg \lambda}+\v v_f\|_2^2
\end{aligned}
\end{equation}

Inverse Kinematics sub-block:
\begin{equation}
\begin{aligned}
    \label{eq:block_2}
    \mathcal{L}_{\rm ik}\revised{(\hat{\vg \phi}, \bar{\vg \phi}, \v v)} = &\sum_{i = 0}^{N} \hspace{2pt} W_p\big\| \mathcal{FK}(\revised{\hat{\v q}}[i])-\mathbf{x}_e^d[i]\big\|_2\\
    &+ \frac{\rho_j}{2}\|\revised{\hat{\v q}} - \revised{\bar{\v q}}+\v v_{\rm ik}\|_2^2
\end{aligned}
\end{equation}

Projection sub-block:
\begin{equation}
\begin{aligned}
    \label{eq:block_3}
    &\mathcal{L}_{\rm proj}\revised{(\vg \phi, \hat{\vg \phi}, \bar{\vg \phi}, \v v)} = \sum_{i = 0}^{N} \hspace{2pt} I_{\mathcal{J},\mathcal{U},\mathcal{F}}(\revised{\Bar{\v q}}[i], \bar{\v u}[i],\Bar{\vg \lambda}[i])\\
    & + \frac{\rho_j}{2}\|\revised{\hat{\v q}} - \revised{\bar{\v q}}+\v v_{\rm ik}\|_2^2
    + \frac{\rho_j}{2}\|\revised{\v q} - \revised{\bar{\v q}}+\v v_j\|_2^2\\
    &+ \frac{\rho_u}{2}\|\v u-\bar{\v u}+\v v_u\|_2^2
    + \frac{\rho_f}{2}\|\vg \lambda - \bar{\vg \lambda}+\v v_f\|_2^2 
\end{aligned}
\end{equation}
Then for each ADMM iteration $k$, the updating sequence in a scaled form is
\begin{subequations}\label{eq:updates}
    \begin{align}
       \label{eq:primal-a}
        &\vg \phi^{k+1}=\underset{\vg \phi}{\arg\min} \ \mathcal{L}_{\revised{\rm wbd}}\revised{(\vg \phi, \bar{\vg \phi}^{k}, \v v^{k})}  \\\nonumber &\hspace{1.2cm} \text{s.t. Eq. }(\revised{\rm \ref{eq:dynamics_constraint_consensus}})\\\label{eq:primal-b} 
        &\hat{\vg \phi}^{k+1}=\underset{\vg \phi_{\rm ik}}{\arg\min} \ \mathcal{L}_{\rm ik}\revised{(\hat{\vg \phi}, \bar{\vg \phi}^{k}, \v v^{k})}\\\label{eq:primal-c}
        &\bar{\vg \phi}^{k+1}=\underset{\bar{\vg \phi}}{\arg\min} \ \mathcal{L}_{\rm proj}\revised{(\vg \phi^{k+1}, \hat{\vg \phi}^{k+1}, \bar{\vg \phi}, \v v^{k})}\\
        &\v v_{\rm ik}^{k+1}=\v v_{\rm ik}^{k}+\revised{\hat{\v q}}^{k+1}-\revised{\Bar{\v q}}^{k+1}\\
        &\v v_j^{k+1}=\v v_j^{k}+\revised{\v q}^{k+1}-\revised{\Bar{\v q}}^{k+1}\\
        &\v v_u^{k+1}=\v v_u^{k}+\v u^{k+1}-\Bar{\v u}^{k+1}\\
        &\v v_f^{k+1}=\v v_f^{k}+\vg \lambda^{k+1}-\Bar{\vg \lambda}^{k+1}
    \end{align}
\end{subequations}
\revised{It is worth noting that both Eq.~(\ref{eq:primal-a}) and Eq.~(\ref{eq:primal-a}) only rely on the solutions from the previous ADMM iteration $k$, which means that the above two steps can be performed in parallel compared with the sequential update shown in Appendix \ref{appendix:admm}.}

To efficiently solve the constrained optimization problem in (\ref{eq:primal-a}), DDP is deployed and the state trajectory is always dynamically feasible by performing the forward pass. For Eq.~(\ref{eq:primal-c}), this minimization problem reduces to a projection operator on convex sets $\mathcal{J}$, $\mathcal{U}$, and $\mathcal{F}$
\begin{equation}
    \begin{aligned}\nonumber
       \bar{\vg \phi}^{k+1}=&\underset{\bar{\vg \phi}\in C}{\arg\min} \ 
         \frac{\rho_j}{2}\|\revised{\hat{\v q}}^{k+1} - \revised{\bar{\v q}}+\v v_{\rm ik}^k\|_2^2 \\ &+ \frac{\rho_j}{2}\|\revised{\v q}^{k+1}-\revised{\Bar{\v q}}+\v v_j^k\|_2^2\\&+ \frac{\rho_u}{2}\|\v u^{k+1}-\Bar{\v u}+\v v_u^k\|_2^2 + \frac{\rho_f}{2}\|\vg \lambda^{k+1}-\Bar{\vg \lambda}+\v v_f^k\|_2^2\\
         C=&\{(\revised{\Bar{\v q}},\bar{\v u},\Bar{\vg \lambda})|\revised{\Bar{\v q}} \in \mathcal{J}, \bar{\v u} \in \mathcal{U},\Bar{\vg \lambda} \in  \mathcal{F}\}
    \end{aligned}
\end{equation}
Then a saturation function can be used to efficiently project the infeasible values onto the boundaries induced by different constraints:
\begin{equation}
\begin{aligned}
    \bar{\vg \phi}^{k+1}=\Pi_{\mathcal{J},\mathcal{U},\mathcal{F}}\Big[&\frac{1}{2}(\revised{\hat{\v q}}^{k+1}+\v v^{k}_{\rm ik}+\revised{\v q}^{k+1}+\v v^{k}_j),\\&
    \v u^{k+1}+\v v_u^k,
    \vg \lambda^{k+1}+\v v_f^k\Big]
\end{aligned}
\end{equation}
The whole process of our ADMM algorithm is shown in \revised{Algorithm \ref{pseudo:DDP-ADMM}}. The selection of $\bar{\vg \phi}$ and dual variables $\v v$ are arbitrary, and we initialized to be zero. The initial trajectory of $\vg \phi$ is generated by running forward dynamics with an initial guess of controls. \revised{The functions $\textit{DDP}(\cdot)$ and $\textit{IK}(\cdot)$ correspond to the implementations of our DDP solver and IK solver.} In each ADMM iteration, the controls from last ADMM iteration will be sent to the current DDP solver as a warm-start, which makes the DDP solver converge faster within around ten iterations in each ADMM iteration after the initial one. Then the trajectories are solved iteratively until a stopping criterion with regard to primal residuals (see \cite{boyd2011distributed}, Sec.~3.3) is satisfied (residuals of magnitude $10^{-2}$).
\begin{algorithm}[t]
  \caption{ADMM trajectory optimization}
  \label{pseudo:DDP-ADMM}
  \begin{algorithmic}[1]
   \STATE $\vg \phi \gets \vg \phi^{0},\hat{\vg \phi} \gets \hat{\vg \phi^{0}}, \bar{\vg \phi} \gets \bar{\vg \phi}^{0}$
   
   \STATE $\v v_j \gets \v v_j^0,\v v_{\rm ik} \gets \v v_{\rm ik}^0, \v v_u \gets \v v_u^0,\v v_f \gets \v v_f^0$
   
    \REPEAT
    \STATE $\vg \phi \gets \text{DDP}\;(\vg \phi,\revised{\bar{\v q}} - \v v_{j},\bar{\v u}-\v v_{u},\bar{\vg \lambda}-\v v_{f})$ \COMMENT{\text{Eq.}~\ref{eq:primal-a}}
    \STATE $\hat{\vg \phi} \gets \text{IK}\;(\revised{\bar{\v q}} - \v v_{\rm ik})$ \COMMENT {\text{Eq.}~\ref{eq:primal-b}}
    \STATE $\bar{\vg \phi} \gets \text{Projection}\;(\frac{1}{2}(\revised{\hat{\v q}}+\v v_{\rm ik}+\revised{\v q}+\v v_j),\v u+\v v_u,\vg \lambda+\v v_f)$ \COMMENT {\text{Eq.}~\ref{eq:primal-c}}
    \STATE $\v v_{\rm ik} \gets \v v_{\rm ik} + \revised{\hat{\v q}} - \revised{\bar{\v q}}$
    \STATE $\v v_j \gets \v v_j + \revised{\v q} - \revised{\bar{\v q}}$
    \STATE $\v v_u \gets \v v_u + \v u - \bar{\v u}$
    \STATE $\v v_f \gets \v v_f + \vg \lambda - \bar{\vg \lambda}$
    \UNTIL{$\rm{stopping\ criterion\ is\ satisfied}$}
    \RETURN{$\vg \phi$}
  \end{algorithmic}
\end{algorithm}

%% file: methods_mpc.tex
\section{Model Predictive Control} \label{section:MPC}
We implement the ADMM planner in a Model Predictive Control (MPC) fashion in the real-time deployment. At each MPC cycle, the optimization in Eq.~(\ref{eq:prime_objective}) is solved with a horizon
$H = N_{\rm steps}\delta t$ and its solution is used as the warm-start for the next MPC cycle, i.e. $(\v x_0, \v u_0) \xleftarrow{} (\v x[i],\v u[i])$ where $\v x_0$ 
is replaced
by some initial state $\v x[i]$. 
The control input to the robot is computed as below.
\begin{align}
    \v u[i]^* &= \v u[i] + PD(\v x[i], \v x^{\rm curr}) \nonumber\\
    &= \overbrace{\v u[i]}^{ff} + \overbrace{\v K(\v q^{\rm curr}-\v q[i])}^{fb} +  \overbrace{ \delta \v u_{FC}(\v F_{e}[i], \v {F}_e^{\rm curr})}^{fc}
\end{align}
where, $\v x^{\rm curr} =  [\v q^{\rm curr} \hspace{4pt} \dot{\v {q}}^{\rm curr} \hspace{4pt} \v {F}_e^{\rm curr}]^T$, is the filtered (low-pass) current state and $\v K \in \R^{n\times n}$ is a gain matrix. $\v u[i]$ is composed of three terms; the feed-forward($ff$), feed-back($fb$), and the admittance force controller($fc$). 
\remark
The low-level controller in the KUKA Sunrise software\cite{KukaManual} uses user-defined joint space impedance control internally. We set a safe () robot impedance to execute our commands as well as to allow the manipulator to behave safely in the compliant environment. 

The high-level MPC loop is implemented at $5$ Hz with a horizon of $1$ s and a time step of $dt = 0.02$ s. In each MPC iteration, we solve for a trajectory by the ADMM planner as described in Section \ref{sec:admm_opt} and we use C++11 in all of our code implementations\footnote{The code implementation can be found at \url{https://github.com/lasithagt/admm}}. To maintain the contact force accuracy and to avoid instabilities resulting from low frequency control update rate\cite{Kazerooni1989OnControl, Wijayarathne2020IdentificationSurface}, we use a low-level force controller which runs at $100$ Hz where the update rate is appropriate for the compliant (low-stiffness) environment. \par
In one of the ADMM planner blocks, we use DDP to solve for the robot and contact dynamics, which consumes most of the computational power. To make the real-time implementation feasible, we use automatic differentiation for derivatives provided by CppADCodeGen\cite{Giftthaler2017AutomaticEstimation}, and RobCoGen\cite{Frigerio2016RobCoGenLanguages} is used to derive the analytical rigid body dynamic model. On a Linux machine (Intel i$7$) with $3.4$ GHz clock speed, average computation time for a horizon of $1s$ took $150$ ms. Without AutoDiff and analytical models, it takes $1300$ ms. With AutoDiff, it is possible to run MPC at a approximate rate of $5$ Hz. \par Each iteration of the DDP takes on average $10$ ms while we limit the number of DDP iterations to $10$ per ADMM cycle to ensure the solver returns an optimal trajectory on time. We run a maximum of $5$ ADMM iterations in each trajectory computation cycle. While the constraint residuals are not guaranteed to reach the same threshold every trajectory iteration, our experimental results demonstrated that $5$ ADMM iterations were sufficient to reduce residuals to the order of $10^{-2}$. 
\begin{figure}[t] 
\centering
\includegraphics[width=1\linewidth]{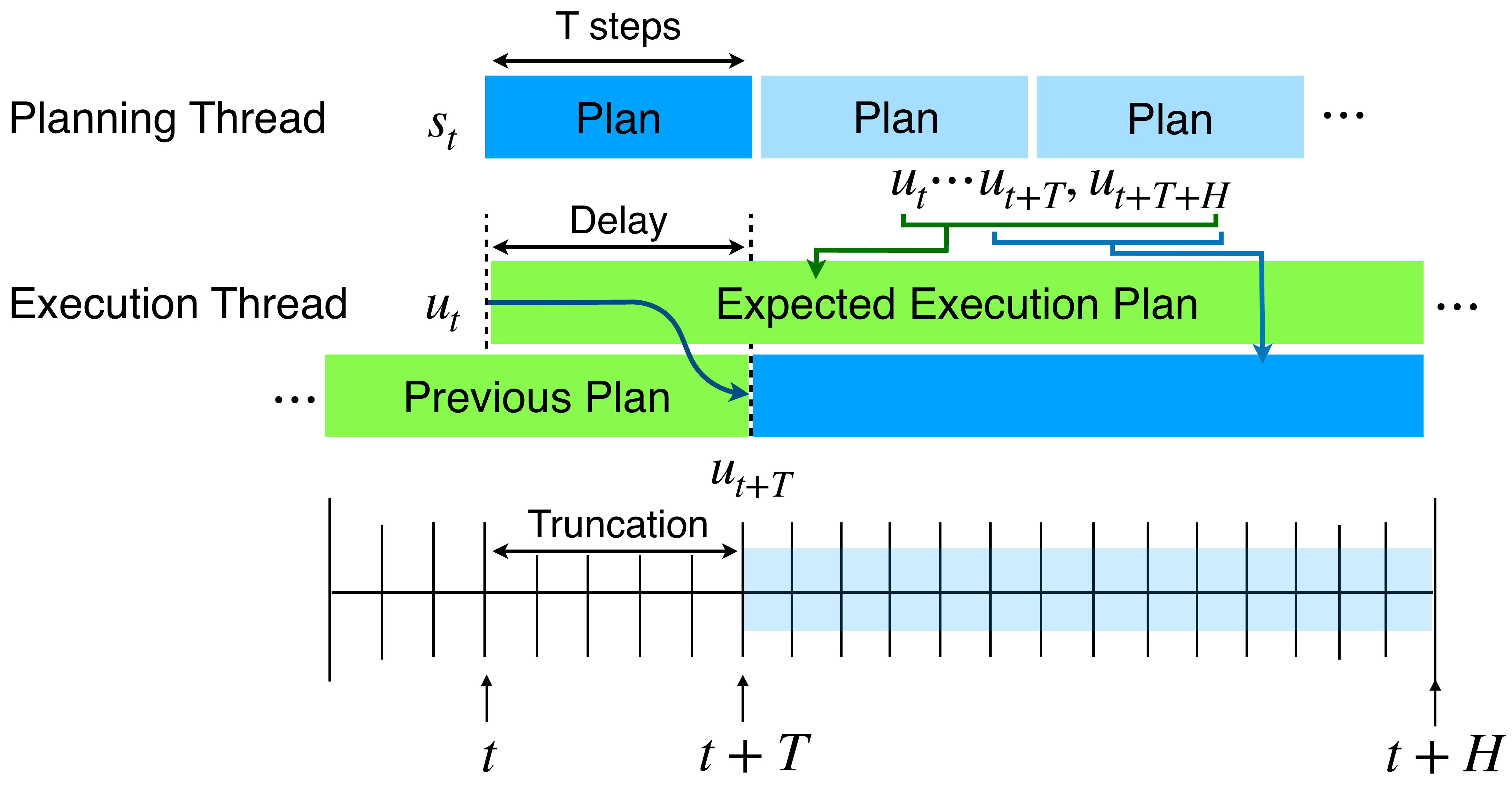}
\caption{Model Predictive Control (MPC) Implementation. $H$, $T$ and $\v s_t$ are the number of timesteps for the horizon, timesteps for the compute delay and current state from the system. Computation takes $T$ steps. During the time, execution thread keeps executing the previous plan. Once planning thread is finished, the current execution plan is updated; the time-steps corresponding to delay is truncated.}
\label{fig:mpc}
\end{figure}
In an ideal MPC setup, the first control input from the output trajectory is applied to the robot and, the current state is used as the initial state for the MPC computation at the next iteration. However, there could be a significant computation burden for the trajectory planner in a practical setting. To take into account the computation and communication delay in the hardware system, we use an asynchronous MPC similar to \cite{Yang2019DataRobots} as shown in Figure \ref{fig:mpc}. The current state ($\v x[t]$) is applied to the planner. It takes $T$ time-steps to compute the trajectory. During that period, the previous trajectory is executed. After $T$ time steps, trajectory $\v U_{t,t+H}=\{\v u_t,\dots, \v u_{t+H}\}$ is returned by the planner. In the actual execution of the trajectory, we truncate the control input sequence up to $T$ time-steps as shown in Figure~\ref{fig:mpc}. The planner and the execution threads are run in parallel in different threads using the threading library in C++11.

Although MPC is beneficial in handling model uncertainties and environmental disturbances, this is not sufficient for tasks that require force modulation. This is due to the instabilities that could arise from contact and determined by the control update rate and surface material properties\cite{Kazerooni1989OnControl, Wijayarathne2020IdentificationSurface}. We use a force controller which updates control input at a rate of $100$ Hz to avoid contact-induced instabilities. Admittance control is proven\cite{Ott2010UnifiedControl} to be better suited for compliant environments where impedance control is best suited for stiff environments.  \par
Force control is important as the surface parameters are not uniform and prone to un-modelled dynamics (e.g., damping, restitution, slipping). To compensate for it, we use admittance control as the force controller in the low-level control. To mitigate the instabilities that would arise from position-based admittance control\cite{Newman1992StabilityControllers}, we use torque as the control input as opposed to position control. The stiffness of the contact surface material was estimated as described in \cite{Wijayarathne2020IdentificationSurface}.

\begin{equation}
    \vg \delta \v u_{FC}[t] =  \v C(x,y,z) \hspace{2pt} \v J^T(\v F_{e}[t]-\v {F}_e^{\rm curr}[t])
\end{equation}
where $\v J(\v q$) is the kinematic jacobian matrix, $\vg \delta \v u_{FC}[t]$ is joint-space torque and $\v C(x,y,z)$ is the compliance matrix which can vary spatially (in the space $(x,y,z)$\footnote{$\v C$ is set to be a constant in this study and identified experimentally}.

%% file: experiments.tex
\section{Experiments}\label{sec:experiments}
\begin{figure*}[t] 
\centering
\includegraphics[width=0.8\linewidth]{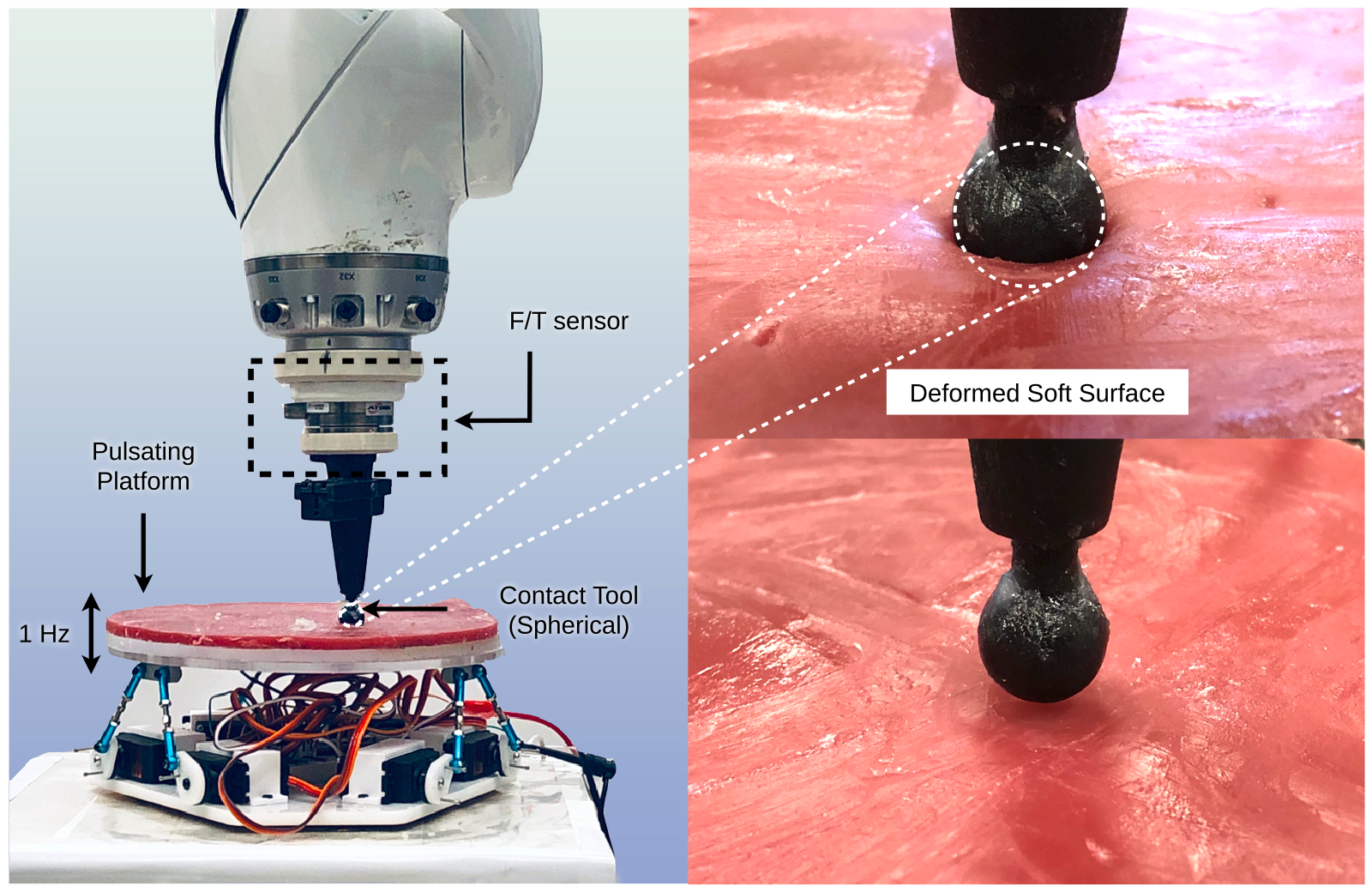}
\caption{Experimental Setup. KUKA manipulator is equipped with spherical tool and a ATI force/torque sensor distally to the contact point. The pulsation platform moves up and down at a fixed rate of $1 Hz$. Surface is covered with a thick soft substrate. The tool of the end-effector is comprised of a spherical indenter which is in contact with the deformable surface.}
\label{fig:experimental setup}
\end{figure*}

To validate the applicability of theoretical attributes of our work in a practical setting, we demonstrate it via physical experiments on a custom-designed platform, shown in Figure \ref{fig:experimental setup}. Moreover, we compare our framework with other state-of-the-art methods which are used widely in robotics motion planning in simulation (in MATLAB$^\copyright$). Then, we show the contact parameter identification methods and compare results with the disturbance-induced tracking task.
\label{experimental validation}

\subsection{Trajectory Optimization Algorithm Comparisons}\label{subsec:TO_benchmark}
The primary motivation for a distributed motion planning scheme such as ADMM is to use different optimization methods that specialize for each sub-problem and incorporate constraints into it. For example, it is efficient to use first-order differential methods to generate IK solutions and solve for the dynamic model separately to find a consensus between them. These were benchmarked in MATLAB$^\copyright$ with a $3.4$ GHz i$7$-core processor. In addition, we benchmark our method against other standard methods, namely:
\begin{enumerate}
    \item \textbf{Our Method:} ADMM with 3-block architecture (Concensus). \textit{The ADMM architecture with three blocks: nonlinear dynamics, IK and projection blocks are solved and concensus is found in the ADMM update. More details can be found in Appendix~\ref{appendix:admm}}
    \item Sequential Quadratic Programming (SQP).
    \textit{Direct collocation with trapizoidal transcription were used with the dynamical system with contact dynamics;} 
    \item Iterative Linear Quadratic Regulator (iLQR - vanilla DDP). \textit{iLQR\cite{li2004iterative} was used with a combined cost of the desired cartesian trajectory ($SE(3)$), force trajectory, state, control, and contact constraints;}
    \item ADMM with 2-block architecture (Sequential).
    \textit{ADMM scheme with 2 blocks as implemented in  \cite{wijayarathne2020simultaneous} was used. In the nonlinear dynamics block, desired state and control cost is used in the DDP solver. Projection block projects to state, control and contact constraints;}
    \item ADMM with 3-block architecture (Sequential).
    \textit{The ADMM architecture with three blocks: nonlinear dynamics, IK and projection blocks are solved sequentially in the ADMM update. More details can be found in Appendix~\ref{appendix:admm}.}
\end{enumerate}

\begin{figure*}[] 
\centering
\includegraphics[width=0.9\textwidth]{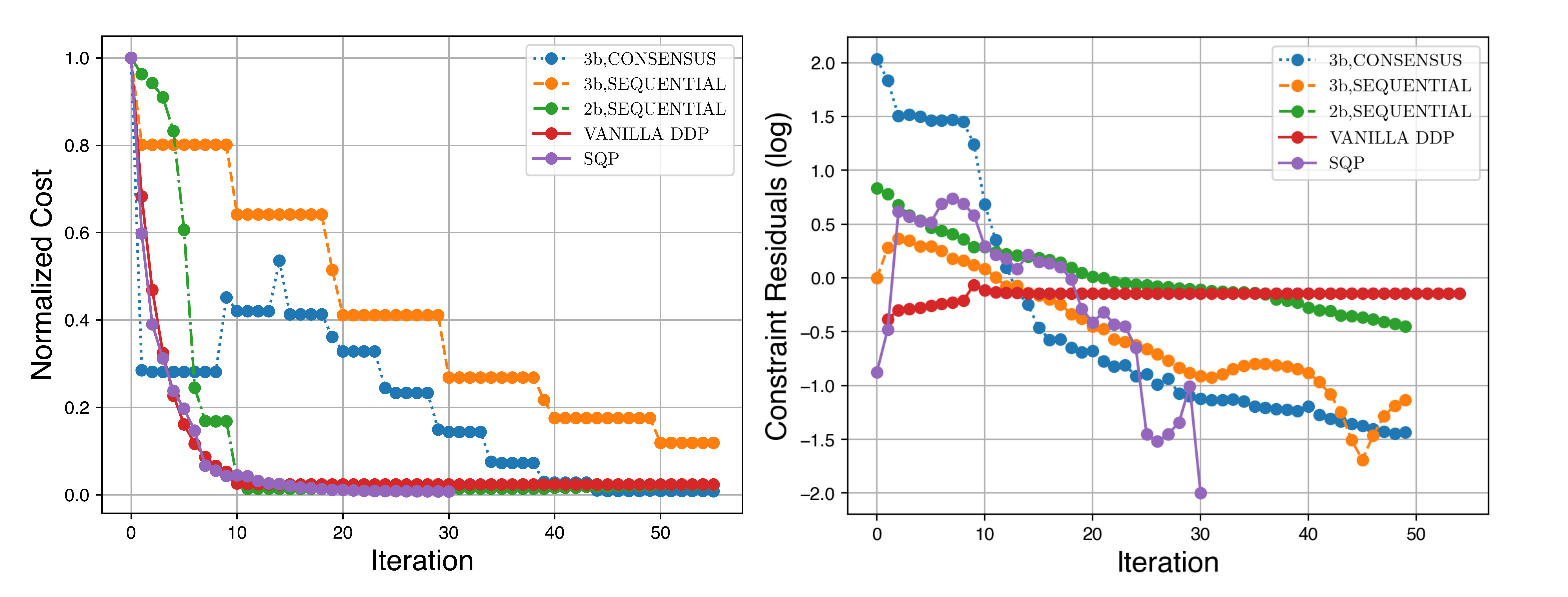}
\caption{Comparison of optimization methods. 3-block consensus, 3-block sequential, 2-block sequential, vanilla-DDP, SQP. \textit{left:} Cost vs Iterations. The cost is normalized such that the initial value is scaled to be unity. \revised{Vanilla DDP and SQP convergence are comparable and 3-block consensus is the fastest.} \textit{right:} Constraint Residuals vs Iterations. \revised{3-block consensus residual convergence is faster although the initial constraint violation is high.}}
\label{fig:admm_comparisons}
\end{figure*}

\begin{table}\centering
\ra{1.3}
\begin{tabular}{c c}
\hline
\textbf{Method} & \textbf{Time(s)} $[10^3]$\\
\hline
3-block CONSENSUS & 3.0966 \\
3-block SEQUENTIAL & 6.6121\\ 
2-block SEQUENTIAL & 16.620\\ 
VANILLA DDP & 4.8436\\ 
SQP & 6.0421\\ \hline
\end{tabular}
\caption{Algorithmic times for 3-block CONSENSUS, 3-block SEQUENTIAL, 2-block SEQUENTIAL, VANILLA DDP and SQP. The whole trajectory was solved with constraints with equal number of time steps. 3-block CONSENSUS takes the shortest time while 2-block SEQUENTIAL takes the longest time.}
\label{table:algorithm_time}
\end{table}

Figure \ref{fig:admm_comparisons} shows the comparison of cost reduction and the contact residuals reduction for the methods described listed above. For the ADMM variants, $x$-axis represents the number of dynamic solver (DDP) iterations instead of ADMM iterations for a fair comparison with other methods (e.g., SQP and iLQR).

\revised{
\subsection{Computation Time Benchmarking}
Table \ref{table:algorithm_time} summarizes the computation time for each algorithm. The algorithms are implemented offline in Matlab for the whole trajectory and derivatives taken with finite differences. It is observed that 3-block CONSENSUS takes the shortest time while 2-block SEQUENTIAL takes the longest time. While the cost and residual convergence of SQP are comparable with 3-block CONSENSUS, it takes about twice the time as 3-block CONSENSUS.} \revisedtwo{Vanilla DDP where the cost is lumped altogether has a comparable computation time to 3-block CONSENSUS but the residual cost does not decay over time as fast as other methods. Overall, the 3-block CONSENSUS method renders the best computation time with a fast residual error decay rate.}
\subsection{Cost Reduction}
One advantage of using ADMM with a 3-block architecture is to track a reference Cartesian trajectory ($SE(3)$) without adding an extra cost term in the dynamics block, which could impede the fast convergence of DDP. \revised{In this approach,} the Cartesian trajectory can be solved efficiently with differential IK and redundancy resolution for redundant systems. Moreover, the solution can warm start the dynamics block. \revised{Figure \ref{fig:admm_comparisons} (\textit{left}) is a comparison of our approach with other existing methods.}  It is evident in the \revised{faster} convergence of ADMM (Figure \ref{fig:admm_comparisons})  with 3-block architecture solved in the consensus manner. Sequential variants tend to converge slower with the \revised{appropriate} penalty parameters ($\rho$) used. \revised{On the other hand, the cost of SQP and Vanilla DDP methods converges comparable to the benchmark but more iterations are needed for the constraint residuals to drop to a satisfactory threshold of $10^{-2}$.}

\subsection{Constraint Satisfaction}
Constraint residuals for each iteration in Figure~\ref{fig:admm_comparisons} (right) are recorded, and the cost coefficients are tuned for each method. It is observed that residuals in the consensus ADMM with 3-block architecture drop relatively faster than other methods. \revised{The }SQP method initially started with a lower residual value and increased (and decreased again). This observation can be attributed to the pre-processing phase of the SQP solver to find an initial solution that is constraint satisfied regardless of the objective cost. \revised{The} iLQR keeps the residual constant without significant improvements.  It is due to the cost and constraints combined cost term, prioritizing cost over constraints.

\subsection{Experimental Results}
In our previous work\cite{wijayarathne2020simultaneous}, we demonstrated the validity of the contact model experimentally when the environmental platform is stationary. 
In this work, we extend it to a non-stationary environment where periodic disturbances are applied ($1Hz$ pulsations) as shown in Figure \ref{fig:experimental setup}. Moreover, we demonstrate the feasibility of the real-time deployment of the proposed framework. The experimental results can be summarized to the categories below:
\begin{figure*}[h!] 
\centering
\includegraphics[width=0.8\linewidth]{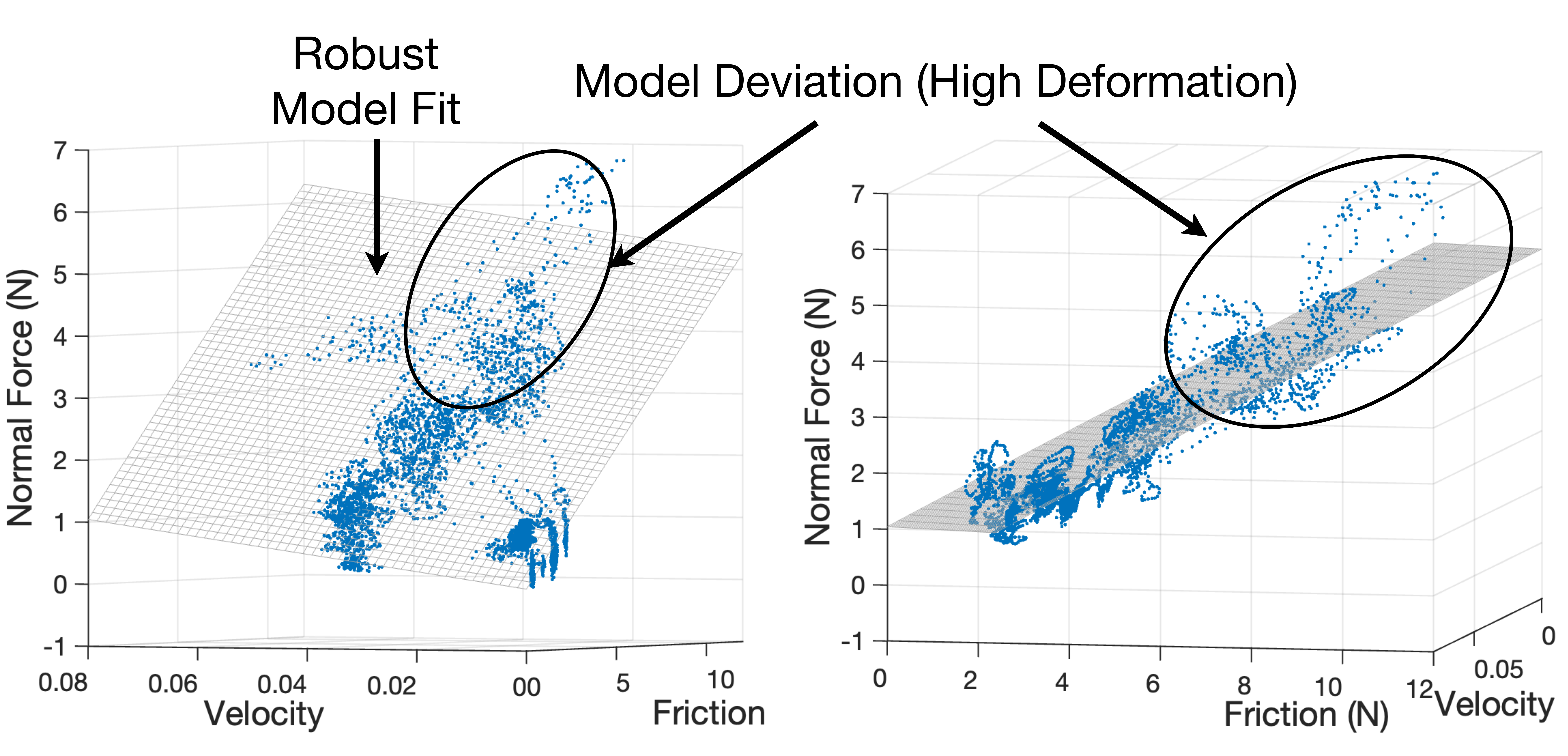}
\caption{Friction model validation and identification. $\mu = 0.4512, K_d = 13.1315, \text{R-squared} = 0.9103$.}
\label{fig:friction_id}
\end{figure*}
\par

\begin{enumerate}
    \item Open-loop trajectories in the presence and absence of environmental disturbances.
    \item Open-loop trajectories with low-level force controller activated.
    \item Model predictive controller (MPC) with low-level force controller in the presence and absence of disturbances.
\end{enumerate}

\subsection{System Identification of Material Properties}
To experimentally validate the proposed soft contact model, parameters related to contact body material need to be identified, e.g., frictional coefficient and Young modulus. The friction coefficient is estimated by performing pre-determined motions along the surface of the material surface while recording the force/torque data through an ATI mini45 sensor which is attached to the end-effector in Figure \ref{fig:experimental setup}.
%

The Young modulus is estimated through performing cyclic linear probing on the surface of the material with the same end-effector point geometry of a sphere (was tested on a material testing platform INSTRON$^\copyright$). It is performed through a non-linear least square estimator by using Eqs.~(\ref{eq:quasi_static_model}) and~(\ref{eq:friction_model}) was used to estimate the frictional coefficients by non-linear least squares estimation as below. \revisedtwo{Eqs. (\ref{eq:contact_parameters}) }summarizes parameter estimation problem.

\begin{subequations}\label{eq:contact_parameters}
\begin{align}\label{eq:local_cost}
    \nonumber \quad &\min_{E}  \sum_{k=0}^{k=N}\|\bar{d}_k - \frac{9F_k^{2}}{16E^{2}R}^{\frac{1}{3}}\|_2 \quad  \text{Derived from Eq.}~(\ref{eq:quasi_static_model})\\
     \quad 
    &\min_{\mu,  k_d}  
    \sum_{k=0}^{k=N}\|\bar{d}_k - \v \mu F_z\left[1+(2\nu-1)\frac{3a^2}{10R^2}\right] \v n_v \\
    & \quad -k_d\v v_e\|_2 \quad \text{Derived from Eq.}~(\ref{eq:friction_model}) 
\end{align}
\end{subequations}

\revisedtwo{Eqs. \ref{eq:contact_parameters} } presents the identification of parameters. Frictional force magnitude in the moving direction $\v F_{\rm fric}$, velocity magnitude $\v v_{e}$, and normal contact force $\v F_{z}$ are calculated from the collected data. A three-dimensional robust least square approximation is fit with a logistic distance function in MATLAB$\copyright$. This fitting is used to mitigate the sensitivity to the model deviation as the deformation increases  (see Figure~\ref{fig:friction_id}). 
Identified Young Modulus and friction coefficient were incorporated into the overall optimization in Eq.~ (\ref{eq:prime_objective}). Desired states to track are the desired end-effector position ($x_e,y_e,z_e$) and the desired normal contact force $F_z$.

The purpose of identifying material properties is two-fold. First, to validate that the used models are well suited and to use in the trajectory optimization framework to generate optimal open-loop trajectories. Friction data fitting results are presented in Figure \ref{fig:friction_id}. Data were fit with a resulting R-squared value of $0.9103$. Frictional coefficient ($\mu = 0.4512$) and damping coefficient ($k_d=13.1315$) were identified. \par
It is observed that with the increase of normal force on the surface, the effects of deformation dominates the frictional forces. This phenomenon is due to the increased rolling friction and material-specific artifacts, e.g., non-uniformity in frictional coefficient and stress distribution. Moreover, the presence of
fluids or any micro-particular particles will increase the non-uniformity.

In the implementation, optimal state trajectories and inputs are generated through the optimization formulated in Eq.~(\ref{eq:prime_objective}). Constraints were satisfied within $~10^{-2}$ residual value violations in both primal and dual stopping criteria. Figure \ref{fig:friction_id} shows that the contact model used is valid for a range of normal forces. Therefore, the desired contact force is maintained within the valid bound of the friction model. 

\begin{figure*}[h] 
\centering
\includegraphics[width=1.0\textwidth]{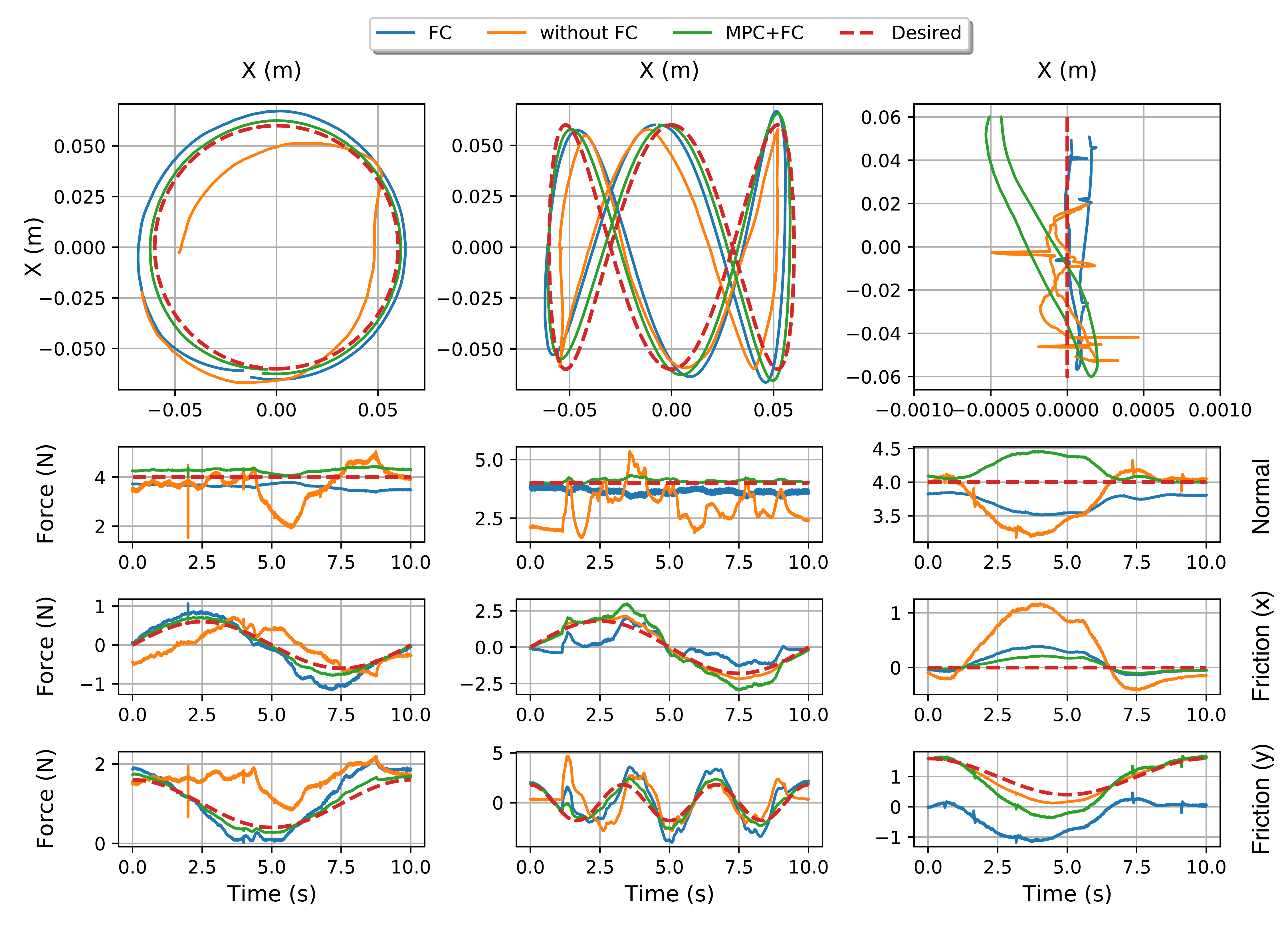}
\caption{\textit{Top row:} Paths tracked by the end-effector vs commanded. \textit{Bottom Row:} Comparison of ground truth of normal and frictional forces vs predicted by the model.}
\label{fig:results_without_pulsations}
\end{figure*}

\subsection{Open-loop Trajectory Generation without Environmental Disturbances}
\label{sec:offline_traj}

\begin{table*}\centering
\ra{1.3}
\begin{tabular}{@{}rrrrcrrcrrrr@{}}\toprule
& \multicolumn{2}{c}{\textbf{Motion Path 1}} & \phantom{abc}& \multicolumn{2}{c}{\textbf{Motion Path 2}} & \phantom{abc}& \multicolumn{2}{c}{\textbf{Motion Path 3}} &
\phantom{abc}\\ \cmidrule{2-4} \cmidrule{5-7} \cmidrule{8-9}
& path (m) & force (N) && path (m) & force (N) && path (m) & force (N)\\ \midrule
Pulsations, without FC & \textbf{0.0142} & 2.723 && \textbf{0.0264} & 2.470 && \textbf{0.0075} & 1.377\\
Pulsations, with FC & 0.0839 & 1.095 && 0.0757 & 1.289 && 0.0187 & 0.8450\\
Pulsations, MPC + FC & 0.0258& \textbf{0.573} && 0.0384 & 0.852 && 0.0153 & \textbf{0.3230}\\ \bottomrule
\end{tabular}
\caption{RSME for motion and force and motion, FC-Admittance Force Control, MPC-Model Predictive Control}
\label{table:offline_results_comparison}
\end{table*}

Open-loop trajectories generated from our trajectory optimization method are used as reference trajectories for the experiments. Any mismatch in contact forces (e.g., those due to friction and deformation) would directly affect the motion and vice versa. It is observed that the control input solved via TO was able to track the motion ($\mathbf{x}_e^d$) and force profile ($\v F_d$) significantly better than a position-controlled robot with force modulation as shown in Figure \ref{fig:results_without_pulsations}. This is due to its open-loop compensation of frictional and centripetal forces encountered during the contact interaction. Table \ref{table:offline_results_comparison} presents the quantification of the results shown in Figure \ref{fig:results_without_pulsations}. \revised{Although the tracking performance is not superior even with the proposed method, it is significantly improved with the existance of model inaccuracy caused by the soft material's unmodelled dynamics, stiction, and non-uniformity. We elaborate on the results in the following paragraphs.}

\begin{figure*}[h] 
\centering
\includegraphics[width=\textwidth]{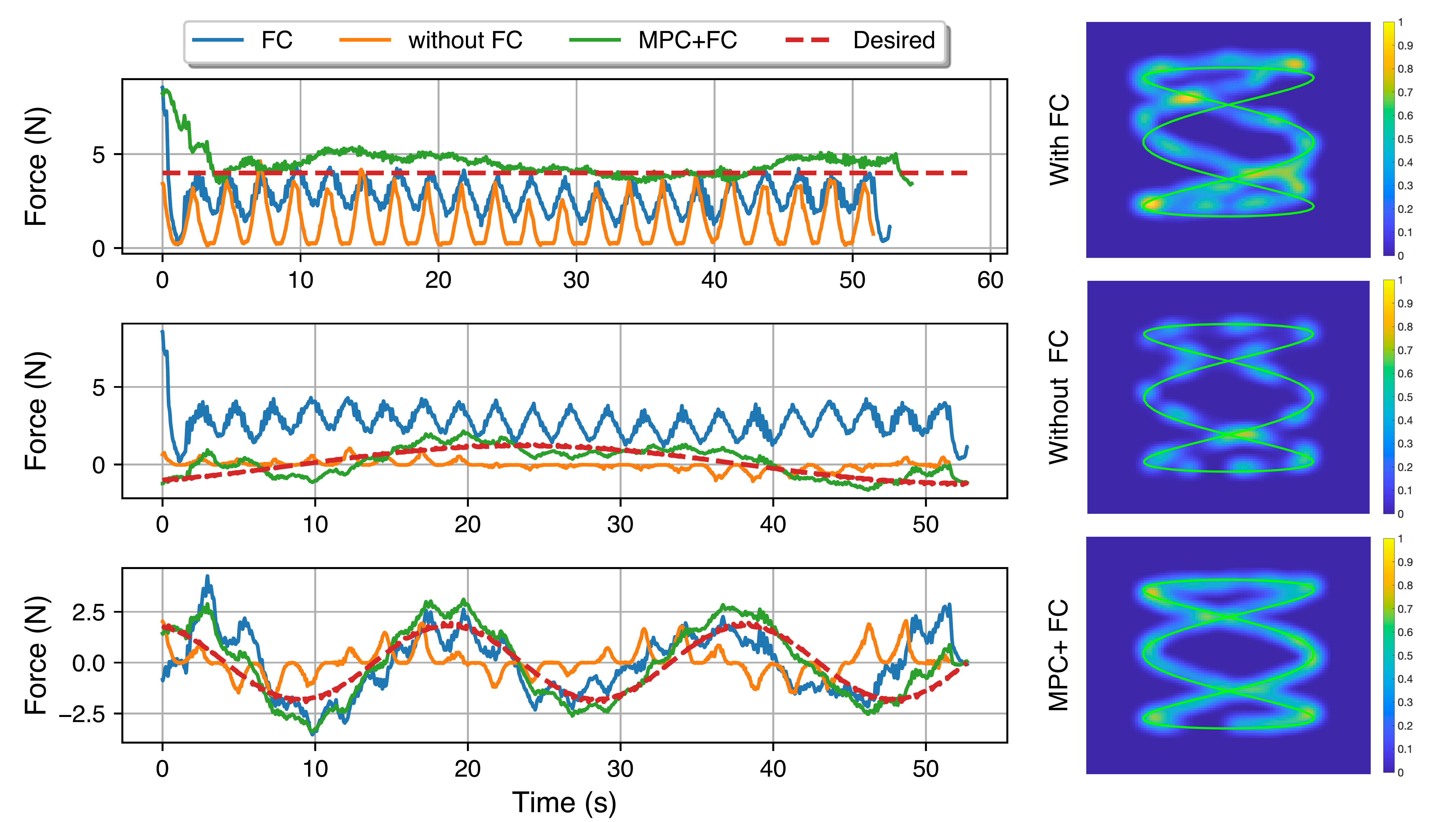}
\caption{\textit{left:} Normal and frictional forces with surface pulsations. Comparison of the activated force controller ($blue$), inactivated force control ($orange$), and MPC and activated force control ($green$). $right$ Normalized force and motion superimposed on a heat map. Note that, motion pulsations of the experimental platform (Figure~\ref{fig:experimental setup}) cause the ripples observed in force profiles, for which the controller is attempting to compensate.}
\label{fig:results_with_pulsations}
\end{figure*}

\begin{table*}\centering
\ra{1.3}
\begin{tabular}{@{}rrrrcrrcrrrr@{}}\toprule
& \multicolumn{2}{c}{\textbf{Motion Path 1}} & \phantom{abc}& \multicolumn{2}{c}{\textbf{Motion Path 2}} & \phantom{abc}& \multicolumn{2}{c}{\textbf{Motion Path 3}} &
\phantom{abc}\\ \cmidrule{2-4} \cmidrule{5-7} \cmidrule{8-9}
& path (m) & force (N) && path (m) & force (N) && path (m) & force (N)\\ \midrule
without FC & 0.284 & 1.042 && 0.392 & 1.328 && 0.302 & 0.891\\
with FC & 0.0183 & 0.384 && 0.0258 & 0.481 && 0.0172 & 0.0273\\
MPC + FC & \textbf{0.0164} & \textbf{0.283} && \textbf{0.0191} & \textbf{0.319} && \textbf{0.0233} & \textbf{0.0084}\\ \bottomrule
\end{tabular}
\caption{RSME for motion and force and motion, FC-Admittance Force Control, MPC-Model Predictive Control, Motion Path 1 (Circle), Motion Path 2 (Eight), Motion Path (Line).}
\label{table:online_results_comparison}
\end{table*}
As described in Section \ref{sec:offline_traj}, the same open-loop full trajectories were executed with force control as described in Section \ref{section:MPC}. Figure \ref{fig:results_without_pulsations} shows experimental results, which suggest low-level force control alone (the blue trajectory) can improve the reference force tracking accuracy significantly. However, the reference motion tracking accuracy degrades due to the frictional forces encountered on the surface observed in the experimental results. The reference force and motion tracking performance improve with active force control and MPC (the green trajectory). Furthermore, Table \ref{table:offline_results_comparison} shows quantified results on multiple motion trajectories with different geometries with different curvatures. \revised{The motion accuracy is observed to be better without force control but the force tracking is worse. On the  other hand, with force control and MPC, both motion and force accuracy are improved as indicated by the corresponding RSME values.} In Cartesian geometries that contain sharp curves (e.g., rectangular geometry), the centripetal force component is an addition to the frictional force in contact force compensation, the velocity at the corners needs to vary to maintain the path and the contact as illustrated in Figure~\ref{fig:path_curvatures}. Failing to compensate for it could result in sliding and deviating from the desired motion (the \textit{without FC} case in Figure \ref{fig:results_without_pulsations}). Only friction force needs to be compensated in a straight line, while both centripetal and frictional components are present in a curved geometry.

\subsection{Model Predictive Control with Low-level Force Controller under Periodic Environmental Disturbances}
\revised{In Section \ref{sec:offline_traj}}, it was shown that force control and MPC could improve both force and motion tracking accuracy. However, the environment is subject to motion disturbances in a realistic setting. To cope with such disturbances and compensate for frictional forces simultaneously, we experimentally show the efficacy of the proposed trajectory optimization (TO) method. Figure \ref{fig:results_with_pulsations} illustrates the force and motion tracking accuracy when under force control, without force control, and with force control, and MPC. Similar to the static case, low-level force control improves force tracking, but the "reactive" nature can be observed in the force tracking (in Figure \ref{fig:results_with_pulsations} right column, second row subplot). As a result, the MPC with contact model information improves both force and motion tracking accuracy significantly.
Moreover, the system remains stable with externally induced motion disturbances and model uncertainties. Such a success can be attributed to the lower-level admittance controller and the model-based TO run in a MPC fashion. Experiments were run on multiple Cartesian trajectories to validate on varying geometries, and quantified results are reported in Table \ref{table:online_results_comparison}.

\subsection{Discussion and Limitations}
The experiments and results demonstrated the importance of incorporating contact in the TO framework for better force and motion tracking accuracy. In safety-critical applications, stiff, position-controlled robots are not desired as they could raise safety concerns. Contact-model-based force-controlled control architectures could provide safe and improved performance. Our static environment results exhibit better performance compared to a motion-induced dynamic environment. Fitting the environment motion disturbance (e.g., breathing, heart-beat) to a parametric model can further improve the tracking performance. In our current work, the robot's impedance is set to a predefined mode that can be limiting. For example, a high impedance robot is more suited for tasks requiring more motion accuracy over force and vice versa. Adapting the robot's impedance depending on tasks and applications can further improve safety, force, and motion tracking accuracy. 
\par
While our work presents a method for force and motion TO, we acknowledge multiple limitations pertaining to the application and computational techniques. For example, the contact model we use is only valid locally. For large deformable bodies, the contact model will require a more computationally expensive method (e.g., Finite Element Methods). Furthermore, the range of force magnitudes was maintained throughout the experiments to be within a local range (in the linear range of Figure~\ref{fig:friction_id}). To compensate for the friction in high deformation and lubrication cases, additional factors of the deformation friction will have to be learned or modeled.

Although this work achieved MPC online planning, there are more promising TO parallelization mechanisms \cite{park2014high, ichter2020perception, plancher2018performance} that can further improve the computational performance. For instance, (i) in the DDP backward pass, a computation speed-up is achievable through a parallelization method of Riccati recursion \cite{farshidian2017real}; (ii) In the DDP forward pass, computation speed-up can be targeted by first proposing an approximate physics model, where a computationally cheap estimate of a coarse model can be evaluated \cite{agboh2019combining}. Then the generated coarse trajectory can be used as an initial seed of the TO with a fine-grained model. This coarse-fine problem can be solved in an iterative procedure. (iii) Finally, these mechanisms above will suit a paralleled ADMM implementation on GPU processors. Although this work does not focus on TO parallelization mechanisms, these potential directions are worth to be reported and can be insightful to the deformable tissue manipulation and medical robotics community.

%% file: conclusion.tex
\section{Conclusion and Future Work}
\label{sec:future work}
In automation tasks requiring physically soft tissue contact, it is paramount to design soft contact interaction models where controllers can be designed to guarantee safety performance. Contact modeling is crucial in correctly identifying the contact material and performing mundane tasks such as incisions along given paths and motion disturbance compensation. This study presented a coherent framework for simultaneous motion and force modulation on compliant surfaces. Moreover, we presented a distributed (ADMM), real-time framework executed in a MPC fashion capable of handling state, control, and contact constraints. Further, we incorporated a soft contact dynamical model into the trajectory optimization (TO). Results proved that motion and force tracking accuracy is significantly improved in both static and dynamic environments. Potential applications of this work include contact manipulation in soft tissues or safety-critical environments.
\par
Trajectories solved from the TO were experimentally validated on a soft surface (EcoFlex$^\copyright$) with the aid of a robot manipulator with an attached spherical shaped tooltip. Surface material properties were estimated and further used in generating optimal trajectories. Experiments were performed on a static and a motion-induced dynamic environment. Results of MPC, with and without force control, were presented. Ground truth forces were obtained using a force-torque sensor (ATI mini45) and compared against the obtained results. MPC with force control was able to track both motion and force both in a static and a dynamic environment with significant improvements. Results and discussion conclude model-based contact modeling and hierarchical TO (e.g., low-level and high level) provide a better alternative for safe simultaneous force and motion generation.
\par 
The future extension of this work is to improve the generability (i.e., "richness") of the contact model to adapt to a wide range of material properties. Furthermore, real-time estimation of the contact model properties can improve the adaptability of the planning framework. Moreover, we intend to extend the work to plan trajectories in three-dimensional surfaces to demonstrate practical applications such as planning robotic incisions on a human body. 

\section{ACKNOWLEDGEMENTS}
The authors would like to thank Qie Sima for his valuable insights and contribution to developing and testing the contact model. 

%% file: appendix.tex
\appendices
\section{Deformable Patch Contact Model}
\label{appendix:deformable_contact}
In this Appendix, we provide the details on several stress distributions. First, the normal Stress Distribution $\sigma_z$:
\begin{equation}
 \frac{\sigma_z}{p_m}=-\frac{3}{2}\left(1-\frac{r^2}{a^2}\right)^{\frac{1}{2}}\quad (r\leq a)
\end{equation}
\\
Radical Stress Distribution $\sigma_r$:
\begin{equation*}
     \frac{\sigma_r}{p_m}=\frac{2\nu-1}{2}\frac{a^2}{r^2}\left[1-\left(1-\frac{r^2}{a^2}\right)\right]-3\nu\left(1-\frac{r^2}{a^2}\right)^{\frac{1}{2}}\quad (r\leq a)
\end{equation*}
Hoop Stress Distribution $\sigma_\theta$:
\begin{equation*}
     \frac{\sigma_\theta}{p_m}=\frac{1-2\nu}{2}\frac{a^2}{r^2}\left[1-\left(1-\frac{r^2}{a^2}\right)\right]-\frac{3}{2}\left(1-\frac{r^2}{a^2}\right)^{\frac{1}{2}}\quad (r\leq a)
\end{equation*}
where $p_m=F/(\pi a^2)$ is the average stress applied in contact part by manipulation and $a=\sqrt{Rd}$ is the radius of contact area (refer to Figure \ref{fig:contact_ball}). The transformation matrix $T$ is 
\begin{equation*}
   T= {\left[ \begin{array}{ccc}
c\theta & s\theta & 0\\
-s\theta &c\theta & 0\\
0 & 0 & 1
\end{array} 
\right ]}
\end{equation*}


\section{Three-block Sequential ADMM}
\label{appendix:admm}
Instead of establishing a consistency constraint between the decision variables from the IK sub-block \revised{and} the projection sub-block \revised{(i.e., $\hat{\v q} = \bar{\v q}$ in Formulation \ref{formulation_admm_consensus})}, the sequential ADMM enforces an equality between DDP sub-block and IK sub-block, \revised{i.e. $\v q = \hat{\v q}$}, as shown in Formulation \ref{formulation_admm_sequential}.
\begin{formulation}
\caption{Distributed Constrained Optimization (Sequential)}\label{formulation_admm_sequential}
\begin{subequations}\label{eq:OCP_three_block}
\begin{align}\nonumber
    &(\textit{\rm{Tracking Task}}) \quad \underset{\vg \phi, \hat{\vg \phi},\bar{\vg \phi}}{\text{min}} \ \sum_{i = 0}^{N} \hspace{2pt}  \delta\mathbf{F}[i]^T\mathbf{Q}_F\hspace{2pt}\delta\mathbf{F}[i]
    + \mathbf{u}[i]^T\mathbf{R}\mathbf{u}[i]  \\ \nonumber 
    &\hspace{3.5cm} + W_p\big\| \mathcal{FK}(\v q[i])-\mathbf{x}_e^d[i]\big\|_2^2 \\
    \nonumber &\hspace{3.5cm} +  I_{\mathcal{J},\mathcal{U},\mathcal{F}}(\Bar{\v q}[i], \bar{\v u}[i],\Bar{\vg \lambda}[i]) \\ \nonumber
    &(\textit{\rm{Variables-DDP}}) \hspace{0.5cm} \vg \phi[i]=[[\overbrace{\v q[i] \hspace{4pt} \dot{\v q}[i] \hspace{4pt} \mathbf{F}_e[i]}^{\v q[i]}]^T, \v u[i]^{T}]^T \\ \nonumber
    &(\textit{\rm{Variables-IK}}) \hspace{0.84cm} \hat{\vg \phi}[i]=\hat{\v q}[i]^T\\ \nonumber
    &(\textit{\rm{Variables-Proj}}) \hspace{0.6cm} \bar{\vg \phi}[i]=[\bar{\v q}[i]^T, \bar{\v u}[i]^T, \bar{\vg \lambda}[i]^T]^T\\
    &\hspace{2.95cm} \forall i = 0, 1, \ldots, N-1 \\\label{eq:dynamics_constraint}
    &(\textit{\rm{Dynamics}}) \hspace{1.57cm} \text{s.t.} \quad 
    \mathbf{x}[i+1] = \mathcal{F}(\mathbf{x}[i], \mathbf{u}[i])\\
    &(\textit{\rm{Initial Condition}}) \hspace{1.4cm}
    \mathbf{x}[0] = \mathbf{x}_0 \\\label{eq:consistency_constraints}
    &(\textit{\rm{Consistency Constraints}}) \hspace{0.2cm} \begin{cases}\v q = \hat{\v q}\\
    \v q = \bar{\v q} \\
    \v u=\bar{\v u} \\
    \vg \lambda = \bar{\vg \lambda}
    \end{cases}
\end{align}
\end{subequations}
\end{formulation}


Different from the consensus variant, the original optimization problem is separated into:

Whole-body dynamics sub-block:
\begin{equation}
\begin{aligned}
    \label{eq:block_1_sequential}&
    \mathcal{L}_{\rm wbd}\revised{(\vg \phi, \hat{\vg \phi}, \bar{\vg \phi}, \v v)} = \sum_{i = 0}^{N} (\delta\mathbf{F}[i]^T\mathbf{Q}_F\hspace{2pt}\delta\mathbf{F}[i]
    + \mathbf{u}[i]^T\mathbf{R}\mathbf{u}[i]) \\
    \nonumber & + \frac{\rho_j}{2}\|\v q - \hat{\v q}+\v v_{\rm ik}^k\|_2^2+ \frac{\rho_j}{2}\|\v q - \bar{\v q}+\v v_j^k\|_2^2 \\
    \nonumber & + \frac{\rho_u}{2}\|\v u-\bar{\v u}+\v v_u\|_2^2 + \frac{\rho_f}{2}\|\vg \lambda - \bar{\vg \lambda}+\v v_f\|_2^2
\end{aligned}
\end{equation}

Inverse kinematics sub-block:
\begin{equation}
\begin{aligned}
    \label{eq:block_2_sequential} &\hspace{-1.8cm}
    \mathcal{L}_{\rm ik}\revised{(\vg \phi, \hat{\vg \phi}, \v v)} = \sum_{i = 0}^{N} W_p\big\| \mathcal{FK}(\hat{\v q}[i])-\mathbf{x}_e^d[i]\big\|_2\\\nonumber
    &\hspace{-1.8cm}+ \frac{\rho_j}{2}\|\v q - \hat{\v q}+\v v_{\rm ik}\|_2^2
\end{aligned}
\end{equation}

Projection sub-block:
\begin{equation}
\begin{aligned}
    \label{eq:block_3_sequential}
    &\hspace{-0.5cm} \mathcal{L}_{\rm proj}\revised{(\vg \phi, \bar{\vg \phi}, \v v)} = \sum_{i = 0}^{N} \hspace{2pt} I_{\mathcal{J},\mathcal{U},\mathcal{F}}(\Bar{\v q}[i], \bar{\v u}[i],\Bar{\vg \lambda}[i])\\
    \nonumber & + \frac{\rho_j}{2}\|\v q - \bar{\v q}+\v v_j\|_2^2 + \frac{\rho_u}{2}\|\v u-\bar{\v u}+\v v_u\|_2^2 \\
    \nonumber &
    + \frac{\rho_f}{2}\|\vg \lambda - \bar{\vg \lambda}+\v v_f\|_2^2 
\end{aligned}
\end{equation}